\documentclass[10pt,journal,compsoc]{IEEEtran}
\ifCLASSOPTIONcompsoc
  \usepackage[nocompress]{cite}
\else
  \usepackage{cite}
\fi
\ifCLASSINFOpdf
  \usepackage[pdftex]{graphicx}
\else
  \usepackage[dvips]{graphicx}
\fi
\usepackage{amsmath, amsfonts}
  \usepackage[caption=false,font=footnotesize,labelfont=sf,textfont=sf]{subfig}
  \usepackage[caption=false,font=footnotesize]{subfig}
\usepackage{tikz}
\newcommand\copyrighttextfinal{%
\scriptsize\copyright\ 2025 IEEE. Personal use of this material is permitted. Permission from IEEE must be obtained for all other uses, in any current or future media, including reprinting/republishing this material for advertising or promotional purposes, creating new collective works, for resale or redistribution to servers or lists, or reuse of any copyrighted component of this work in other works.}%
\newcommand\copyrightnotice{%
\begin{tikzpicture}[remember picture,overlay]%
\node[anchor=south,yshift=10pt] at (current page.south) {{\parbox{\dimexpr\textwidth-\fboxsep-\fboxrule\relax}{\copyrighttextfinal}}};%
\end{tikzpicture}%
}

\usepackage{url}

\usepackage{multirow}
\usepackage{booktabs}
\usepackage{makecell}
\usepackage{blindtext}
\usepackage{xcolor}
\usepackage{color, colortbl}
\definecolor{grays}{rgb}{0.94, 0.94, 0.94}
\newcolumntype{C}[1]{>{\centering\arraybackslash}m{#1}}
\newcolumntype{L}[1]{>{\raggedright\arraybackslash}m{#1}}
\newcolumntype{+}{@{\hskip\tabcolsep\vrule width -2pt\hskip\tabcolsep}}

\begin{document}
%
\title{Revisiting Gradient-based Uncertainty for Monocular Depth Estimation}
%
%
%
%

\author{Julia Hornauer, Amir El-Ghoussani and Vasileios Belagiannis
\IEEEcompsocitemizethanks{\IEEEcompsocthanksitem Julia Hornauer is with Ulm University, Ulm, Germany.\protect\\
E-mail: julia.hornauer@alumni.uni-ulm.de
\IEEEcompsocthanksitem Vasileios Belagiannis and Amir El-Ghoussani are with Friedrich-Alexander-Universität Erlangen-Nürnberg, Erlangen, Germany.\protect\\
E-Mail: \{firstname\}.\{lastname\}@fau.de}
\thanks{Manuscript received Month 07, 2023; revised Month 01, 2025.}}

\IEEEtitleabstractindextext{%
\begin{abstract}
Monocular depth estimation, similar to other image-based tasks, is prone to erroneous predictions due to ambiguities in the image, for example, caused by dynamic objects or shadows.
For this reason, pixel-wise uncertainty assessment is required for safety-critical applications to highlight the areas where the prediction is unreliable. 
We address this in a post hoc manner and introduce gradient-based uncertainty estimation for already trained depth estimation models. 
To extract gradients without depending on the ground truth depth, we introduce an auxiliary loss function based on the consistency of the predicted depth and a reference depth. 
The reference depth, which acts as pseudo ground truth, is in fact generated using a simple image or feature augmentation, making our approach simple and effective. 
To obtain the final uncertainty score, the derivatives \textit{w.r.t.} the feature maps from single or multiple layers are calculated using back-propagation. 
We demonstrate that our gradient-based approach is effective in determining the uncertainty without re-training using the two standard depth estimation benchmarks KITTI and NYU. In particular, for models trained with monocular sequences and therefore most prone to uncertainty, our method outperforms related approaches. In addition, we publicly provide our code and models: \url{https://github.com/jhornauer/GrUMoDepth}.
\end{abstract}

}

\maketitle

\IEEEdisplaynontitleabstractindextext

%
\IEEEpeerreviewmaketitle

\begin{figure}
    \centering
    \includegraphics[width=0.93\linewidth]{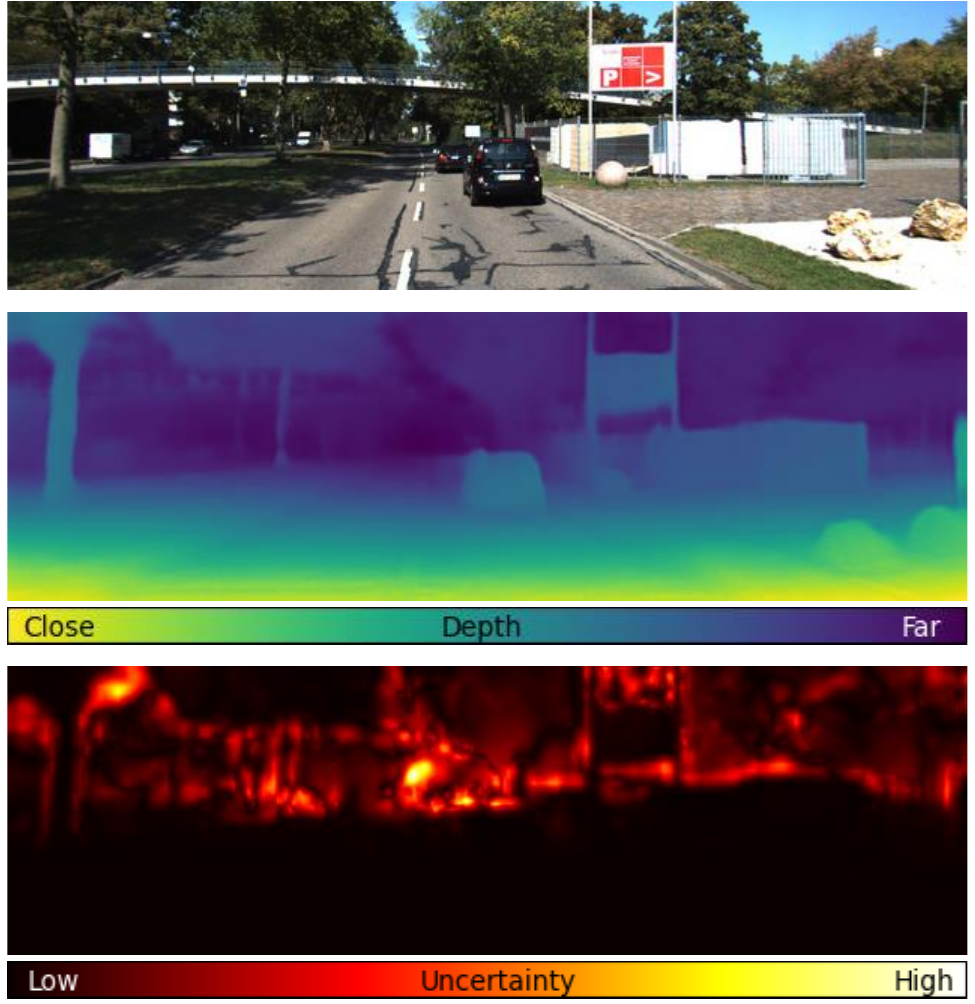}
    \caption{Example image from KITTI~\cite{Geiger2013IJRR} with corresponding depth prediction and uncertainty estimated with our gradient-based method. The uncertainty estimate demonstrates that the predicted depth is not reliable in the region where shadows and occlusions appear.}
    \label{fig:teaser}
\end{figure}

\IEEEraisesectionheading{\section{Introduction}\label{sec:introduction}}
\IEEEPARstart{3}{D} perception such as depth estimation is important for environment modelling in safety-critical applications like automated driving or robotics. Recent advances in deep neural networks favour monocular depth estimation, which is advantageous due to the cheaper availability of cameras compared to LiDAR sensors while providing higher resolution and frame rates. However, similar to other image-based tasks, ambiguities due to, for example, occlusions or shadows lead to incorrect depth predictions. In Fig.~\ref{fig:teaser}, big trees around the bridge make depth perception difficult, leading to wrong depth estimates. To identify potentially erroneous regions in the prediction, it is essential to perform an uncertainty assessment along with the depth prediction.  

\IEEEpubidadjcol
Recent approaches address uncertainty estimation for monocular depth estimation with log-likelihood maximisation by placing a distribution over the model output~\cite{Poggi2020OnTU,Klodt2018SupervisingTN}. 
Most approaches require a change in the training procedure~\cite{Poggi2020OnTU,amini2020deep,Klodt2018SupervisingTN}, wherefore the model must be trained from scratch and the focus shifts to uncertainty estimation, which can affect depth estimation performance. However, re-training is not always feasible if, for example, the model is adapted to the target hardware, e.g., by pruning, or the model is provided externally and the training data is not accessible. 

Alternative approaches comprise bootstrapped ensembles~\cite{Lakshminarayanan2017SimpleAS} or Monte Carlo Dropout~\cite{mcdropout}. These methods require extensive sampling to obtain an accurate uncertainty estimate, which does not meet the real-time requirements of safety-critical applications. For this reason, we address uncertainty estimation for already trained depth estimation models, which makes our method a \textit{post hoc} approach.
\copyrightnotice

Previous approaches realise \textit{post hoc} uncertainty estimation by training a second model~\cite{Upadhyay2022BayesCapBI,hornauer2023out} or sampling during inference~\cite{Mi2019TrainingFreeUE} with augmentation or dropout applied. In contrast, we present a computationally lighter approach by extracting gradients from the already trained model using an auxiliary objective, which only requires an additional backward pass. Therefore, our method incurs no additional training overhead and comparatively little additional effort during inference, while providing consistent uncertainty estimation performance across several depth estimation scenarios with different models and datasets. 

While gradients are mostly used to learn neural network parameters, recent works improve the robustness of classification models by using gradients to detect unknown inputs~\cite{Oberdiek2018ClassificationUO,Huang2021OnTI}. Inspired by these gradient-based approaches, we propose to estimate the pixel-wise uncertainty of the depth estimation model predictions using gradients extracted from the trained and therefore fixed model. To extract gradients that adequately reflect predictive uncertainty, we introduce an auxiliary loss function that is independent of the ground truth depth. To this end, we construct a reference depth by augmenting the image or feature space and passing the augmented version through the fixed neural network. The auxiliary loss, which reflects the inconsistency between the predicted depth and the reference depth, is then back-propagated through the depth estimation model. The computed gradients \textit{w.r.t} the decoder feature maps are finally used to calculate the uncertainty estimation score. We consider either a single layer or multiple decoder layers to calculate the uncertainty score. Both options lead to state-of-the-art uncertainty estimation results on two standard depth estimation benchmarks, namely KITTI~\cite{Geiger2013IJRR} and NYU Depth V2~\cite{SilbermanECCV12}, without model training or costly sampling.   

This work extends our earlier conference publication on gradient-based uncertainty for
monocular depth estimation~\cite{Hornauer2022GradientbasedUF} with improvements in the following aspects: 1) We enhance the gradient-based uncertainty estimation to consider not only image-based augmentation to generate the reference depth but also feature-based augmentation. 2) In addition, we improve the uncertainty score calculation by introducing the technique of calculating the uncertainty score from multiple layers instead of a single layer, which eliminates the need for careful layer selection. 3) We demonstrate the effectiveness of gradient-based uncertainty estimation by including experiments using a recently published transformer-based depth estimation model, showing that our method is independent of the underlying model architecture. For this purpose, we use the normalised uncertainty calibration error (nUCE) as a modification of the UCE~\cite{Laves2019WellcalibratedMU} to evaluate uncertainty in regression tasks.  4) We provide extensive experiments that give further insight into our method by ablating the augmentation used to generate the reference depth, the layer selection, the auxiliary loss function for predictive models, and how to combine gradients from multiple layers. 

\section{Related Work}
In this section, we first provide an overview of existing methods for uncertainty estimation, especially for monocular depth estimation. Then, we discuss several methods that use gradients to analyse the robustness of neural networks. Finally, we present recent work on monocular depth estimation.
\subsection{Uncertainty Estimation}
Neural network uncertainties are mainly categorised into epistemic and aleatoric uncertainties~\cite{Kendall2017WhatUD}. While epistemic uncertainty refers to model uncertainty resulting from lack of knowledge, aleatoric uncertainty refers to data uncertainty caused by noise such as reflections or occlusions~\cite{Kendall2017WhatUD}. The existing approaches to uncertainty estimation can be mainly divided into empirical, predictive, and post hoc methods, depending on how the uncertainty is determined. They deal with either epistemic or aleatoric uncertainty or both, the so-called predictive uncertainty. In the following, we will discuss different methods and the addressed uncertainty type in more detail. 
\subsubsection{Empirical Methods}
Empirical uncertainty estimation methods place a distribution over the model weights and therefore address the epistemic uncertainty. To this end, bootstrapped ensembles~\cite{Lakshminarayanan2017SimpleAS} train multiple models with different initialisation to compute mean and variance over their outputs as prediction and uncertainty measures, respectively. Snapshot ensembles~\cite{HuangSnapshot2017} remove the training overhead by leveraging cyclic learning rate scheduling. Monte Carlo (MC) Dropout~\cite{mcdropout}, on the other hand, applies dropout~\cite{dropout} during training and takes multiple samples with dropout enabled during inference which avoids storing multiple models. 
\subsubsection{Predictive Methods}
In contrast, predictive approaches maximise the log-likelihood by placing a distribution over the model output, which in turn accounts for aleatoric uncertainty. In regression tasks, the Laplacian~\cite{Ilg2018UncertaintyEA} or Gaussian~\cite{Nix1994EstimatingTM} distribution can be used to predict mean and variance as depth estimate and uncertainty measure, respectively. Klodt \textit{et al.}~\cite{Klodt2018SupervisingTN}, transfer this to the photometric loss used in self-supervised monocular depth estimation. Poggi \textit{et al.}~\cite{Poggi2020OnTU} apply a self-learning paradigm in which a second model is trained in a supervised manner with depth predictions from a self-supervised model to learn the depth and corresponding uncertainty prediction by maximising the log-likelihood. 
Instead of targeting only one type of uncertainty, Amini \textit{et al.}~\cite{amini2020deep} start from the evidential distribution to distinguish between aleatoric and epistemic uncertainty.
\subsubsection{Post Hoc Methods}
All of these approaches require a training procedure tailored to uncertainties and thus may affect the depth estimation performance. Therefore, so-called \textit{post hoc} methods estimate the uncertainty of already trained models. Since model re-training is not always desirable, a second model is optimised in~\cite{Upadhyay2022BayesCapBI} assuming the generalised Gaussian distribution to estimate the uncertainty of already trained image-to-image translation models. Similarly, we also estimate the uncertainty of fixed models but without the need to train a second model. Another approach to \textit{post hoc} uncertainty estimation is the approximation of the model output distribution by sampling with input augmentations or dropout applied only during inference~\cite{Mi2019TrainingFreeUE}. In this work, we also estimate the uncertainty of already trained depth estimation models training-free, but without exhaustive sampling strategies, by extracting gradients with an auxiliary loss function. 

\subsection{Model Robustness by Gradient Analysis}
In neural network optimisation, gradients are used as an indication of how to adjust the model weights to learn a mapping function that best represents the given inputs. For this reason, gradients are used in recent works~\cite{Oberdiek2018ClassificationUO,lee2022adversarial,Huang2021OnTI} to determine whether an input is well represented by the model, and thus to detect inputs that are not in the training distribution. For gradient extraction, a loss function must be defined that can be back-propagated through the neural network. While Oberdiek \textit{et al.}~\cite{Oberdiek2018ClassificationUO} use the negative log-likelihood of the predicted class, Lee \textit{et al.}~\cite{lee2022adversarial} utilise the binary cross entropy between the logits and a confounding label defined as a vector containing only ones to identify whether the input can be associated with one of the learned classes. Huang \textit{et al.}~\cite{Huang2021OnTI}, in contrast, rely on the KL divergence between the Uniform distribution and the softmax output. Recently, in~\cite{Riedlinger2021GradientBasedQO}, gradients are used for uncertainty estimation in object detection, whereas Maag and Riedlinger~\cite{Maag2023PixelwiseGU} leverage gradients to segment unknown objects in semantic segmentation. In our work, we use the informativeness of gradients for uncertainty estimation in the computationally expensive dense regression task of monocular depth estimation. In contrast to image classification and object detection, we require a pixel-wise uncertainty score. To accomplish this, we define the auxiliary loss function as the distance between the predicted depth and a reference depth, which makes it independent of ground truth.  

\subsection{Monocular Depth Estimation}
Early works in monocular depth estimation rely on supervised training~\cite{Eigen2014DepthMP,Laina2016DeeperDP,Song2021MonocularDE,Bauer2021NVSMonoDepthIM,Bhat2021AdaBinsDE}. More recently, self-supervision with stereo pairs~\cite{Godard2017UnsupervisedMD,zhou_diffnet,Tosi2019LearningMD} or monocular video sequences~\cite{Zhou2017UnsupervisedLO,monodepth2} is used instead to reduce the need for costly ground truth collection of dense depth maps. 
While most methods rely on a static scene assumption, Bian \textit{et al.}~\cite{Bian2019UnsupervisedSD} and Xu \textit{et al.}~\cite{Xu2019RegionDN} introduce self-discovered masks and deformation-based motion representation to handle dynamic objects, respectively. With the emerging success of vision transforms~\cite{dosovitskiy2020vit,Lee2021MPViTMV}, recent works take advantage of them for monocular depth estimation~\cite{Bhat2021AdaBinsDE,Ranftl2021VisionTF,Yuan2022NeuralWF,Agarwal_2023_WACV,Yang2021TransformerBasedAN}. These works integrate the attention module as backbone~\cite{Ranftl2021VisionTF}, in skip connections~\cite{Agarwal_2023_WACV}, in the decoder~\cite{Yang2021TransformerBasedAN} or combined with Conditional Random Fields~\cite{Yuan2022NeuralWF}. While most works target depth estimation as a regression task, Bhat \textit{et al.}~\cite{Bhat2021AdaBinsDE} phrase it as a classification-regression task. 
In contrast to those supervised trained models, MonoViT~\cite{monovit} is the first transformer-based model trained in a self-supervised manner. 
Different methods aim at zero-shot generalization to unseen inputs by training on a large amount of diverse datasets~\cite{midas,Ranftl2021VisionTF,yang2024depth,yang2024depthv2} or using the representation capabilities of generative models~\cite{ke2024repurposing}.
Therefore, Ranftl~\textit{et al.}~\cite{midas,Ranftl2021VisionTF} propose a robust training objective to handle different depth scales and annotations.
Yang~\textit{et al.}~\cite{yang2024depth} combine training on both labelled and unlabelled data using a teacher-student approach to obtain pseudo-labels for the great amount of unlabelled data. Thereafter, enhancement in finer details is accomplished by replacing the labelled real data with synthetic data~\cite{yang2024depthv2}.
In contrast, Ke~\textit{et al.}~\cite{ke2024repurposing} leverage the representation capabilities of pre-trained diffusion models to obtain improved generalizability in monocular depth estimation only finetuning on sythetic data.
In this work, we propose a method for uncertainty estimation for depth predictions of already trained models. Importantly, our auxiliary loss function makes our approach independent of whether the model is supervised or self-supervised trained. Furthermore, we demonstrate in our experiments that our gradient-based uncertainty is applicable to both convolutional and transformer-based models.

\section{Method}
\begin{figure*}[ht]
    \centering
    \includegraphics[width=0.96\textwidth]{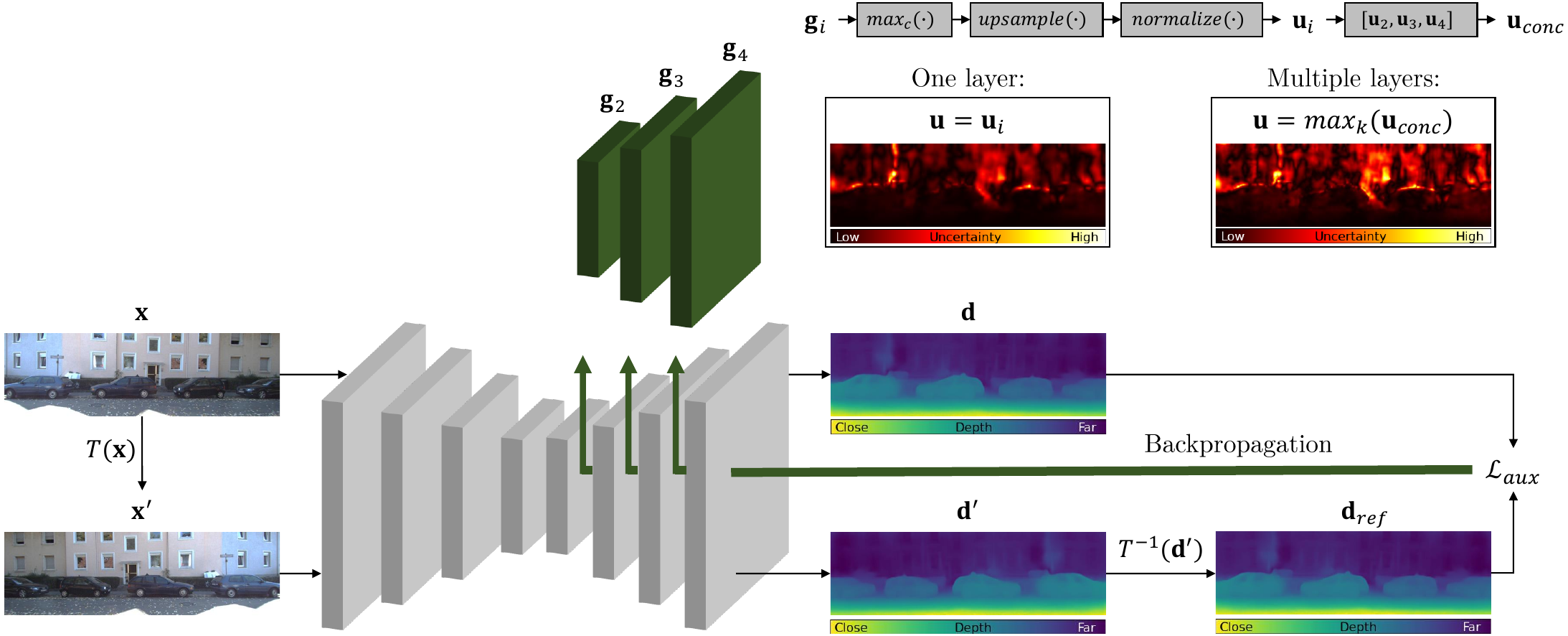}
    \caption{\textbf{Overview of our gradient-based uncertainty estimation method with the horizontal flip as transformation $T(\cdot)$:} First, we apply the transformation $T(\cdot)$ on the image $\mathbf{x}$ to obtain $\mathbf{x}^\prime$. Then, both images are passed through the depth estimation model to obtain the depth estimates $\mathbf{d}$ and $\mathbf{d}^\prime$, respectively. Since $T(\cdot)$ is an invertible geometric transformation, we apply the inverse transformation $T^{-1}(\cdot)$ to the depth estimate $\mathbf{d}^\prime$ to obtain the reference depth $\mathbf{d}_{ref}$. For the gradient extraction, the auxiliary loss $\mathcal{L}_{aux}(\mathbf{d}, \mathbf{d}_{ref})$ is back-propagated through the decoder to extract the gradient maps $\mathbf{g}_{i}$ at different decoder layers $i$. Either one specific layer or multiple layers can be chosen for the gradient extraction. The extracted gradient maps can then be used to calculate the respective uncertainty maps $\mathbf{u}_{i}$. Finally, the pixel-wise uncertainty score $\mathbf{u}$ is chosen to be the uncertainty map obtained from a single uncertainty map $\mathbf{u}_{i}$ or is calculated from $k$ uncertainty maps $\{\mathbf{u}_{i}\}^{k}$.}
    \label{fig:overview}
\end{figure*}

In monocular depth estimation, the goal is to estimate the pixel-wise depth $\mathbf{\hat{d}} \in \mathbb{R}^{w \times h \times 1}$ from an RGB image $\mathbf{x} \in \mathbb{R}^{w \times h \times 3}$, where $w$ and $h$ are the image width and height, respectively. 
We consider a standard depth estimation neural network $f(\cdot)$, parameterised by $\theta$, to predict the pixel-wise depth $\mathbf{\hat{d}} = f(\mathbf{x};\theta)$. Specifically, the model consist of the encoder $\mathbf{z} = \phi(\mathbf{x}; \theta_{\phi})$ with parameters $\theta_{\phi}$ and the depth decoder $\mathbf{\hat{d}} = \psi(\mathbf{z};\theta_{\psi})$ with parameters $\theta_{\psi}$, with the composition given as $\mathbf{\hat{d}}=f(\mathbf{x};\theta)=\psi(\phi(\mathbf{x}; \theta_{\phi});\theta_{\psi})$. In this work, we aim to predict the pixel-wise uncertainty $\mathbf{u} \in \mathbb{R}^{w \times h \times 1}$ corresponding to the depth prediction $\mathbf{\hat{d}}$. For this, we propose a gradient-based uncertainty estimation approach for already trained depth estimation models.\\
\\
\textbf{Gradient-based Uncertainty:}
In particular, we propose to generate the pixel-wise uncertainty scores $\mathbf{u}$ using gradients extracted from the decoder $\psi(\cdot)$ of the depth estimation model. To compensate for the lack of depth ground truth, we introduce a reference depth $\mathbf{d}_{ref}$ based on which we construct an auxiliary loss function. 
We define the auxiliary loss function $\mathcal{L}_{aux}$ at the pixel level for back-propagation through the depth decoder $\psi(\cdot)$ to obtain the derivatives \textit{w.r.t.} the feature representations $\mathbf{a}$.
The resulting gradient maps $\mathbf{g}$ are afterwards used to compute the pixel-wise uncertainty score $\mathbf{u}$. For this purpose, we introduce an approach to obtain the uncertainty score from a single feature representation or multiple feature representations.
We assume that the estimated uncertainty based on the gradients reflects the actual error in the depth estimate. 
Fig.~\ref{fig:overview} gives an overview of our method. Next, we explain how to generate the reference depth and how the auxiliary loss function is defined. Afterwards, we describe the gradient extraction process and how we determine the final pixel-wise uncertainty score. 

\subsection{Reference Depth}\label{sec:refdepth}
Overall, we build our auxiliary loss function $\mathcal{L}_{aux}$ relying on a reference depth $\mathbf{d}_{ref}$ for comparison with the predicted depth $\mathbf{\hat{d}}$. An intuitive choice as a reference would be the ground truth depth $\mathbf{d} \in \mathbb{R}^{w \times h \times 1}$. However, during inference, we do not have access to any ground truth. For this reason, we rely on a reference image $\mathbf{x}^{\prime}$ to generate a reference depth $\mathbf{d}_{ref}$.
The reference image $\mathbf{x}^{\prime} \in \mathbb{R}^{w \times h \times 3}$ is a transformed version of the original input image $\mathbf{x}$, which is obtained by image augmentation. Importantly, the scene structure should be preserved. Therefore, we define the transformation function $T(\cdot)$ to obtain the reference image $\mathbf{x}^{\prime} = T(\mathbf{x})$ from the original input image $\mathbf{x}$. The reference image $\mathbf{x}^{\prime}$ is then forwarded through the depth estimation model $\mathbf{d}^{\prime}=f(\mathbf{x}^{\prime})$ to obtain the corresponding depth prediction $\mathbf{d}^{\prime}$. 
If $T(\cdot)$ is a invertible geometric transformation, the inverse function $T^{-1}(\cdot)$ must be applied to the depth prediction $\mathbf{d}^{\prime}$ such that the two depth estimated $\mathbf{d}^{\prime}$ and $\mathbf{\hat{d}}$ coincide. In this case, the reference depth is obtained by applying the inverse transformation $\mathbf{d}_{ref}=T^{-1}(\mathbf{d}^{\prime})$, otherwise the reference depth is given by the depth prediction $\mathbf{d}_{ref} = \mathbf{d}^{\prime}$. We assume that the reference depth $\mathbf{d}_{ref}$ should match the depth prediction $\mathbf{\hat{d}}$ since $\mathbf{x}^{\prime}$ is either geometrically transformed or slightly modified and thus the pixel information and structure of $\mathbf{x}^{\prime}$ match with the original image $\mathbf{x}$. The difference between the depth prediction $\mathbf{\hat{d}}$ and the reference depth $\mathbf{d}_{ref}$ could already be used to measure uncertainty. 
However, we claim that gradients are more informative for uncertainty estimation compared to only considering the loss at the model output. Our experiments support this claim by showing that uncertainty generated with feature map gradients is more effective in uncertainty estimation than the loss function itself. 

\subsection{Auxiliary Loss Function}\label{sec:auxloss}
After generating the reference depth, we can build the auxiliary loss function $\mathcal{L}_{aux}$ with the predicted depth $\mathbf{\hat{d}}$ and the reference depth $\mathbf{d}_{ref}$. In our work, we consider both regular depth estimation models $\mathbf{\hat{d}} = f(\mathbf{x};\theta)$, which output only the predicted depth, and predictive depth estimation models $\mathbf{\hat{d}}, \mathbf{\sigma} = f_{\sigma}(\mathbf{x};\theta_{\sigma})$ ~\cite{Nix1994EstimatingTM,Ilg2018UncertaintyEA,Poggi2020OnTU}, which output the prediction variance $\mathbf{\sigma}$ as an uncertainty value in addition to the depth estimate. Since predictive depth estimation models already predict the variance, we take this into account for the definition of our auxiliary loss function as presented below. Because our gradient-based approach is a \textit{post hoc} method, it is independent of whether the underlying depth estimation model already predicts an uncertainty value. Our experiments in Section~\ref{sec:exp_results} show that applying gradients to the predictive depth estimation model improves the uncertainty estimated by the model itself.

\subsubsection{Regular Depth Estimation Model}
For regular depth estimation models, we define the auxiliary loss function as the squared difference between the predicted depth and the reference depth: 
\begin{equation}\label{eq:laux}
    \mathcal{L}_{aux} = (\mathbf{\hat{d}} - \mathbf{d}_{ref})^2. 
\end{equation}
Therefore, the gradient extraction is based on the inconsistency between both depth predictions. Since the spatial structure is preserved in convolutions, regions with higher values in the pixel-wise loss lead to larger gradients, while regions with lower values lead to smaller gradients. 

\subsubsection{Predictive Depth Estimation Model}
Predictive depth estimation models $f_{\sigma}$, in contrast, predict the mean $\mathbf{\hat{d}}$ and the variance $\mathbf{\sigma}$ as depth prediction and the uncertainty score, respectively. This is accomplished by assuming a distribution over the model output. When optimising the $\mathcal{L}_{1}$ or $\mathcal{L}_{2}$ loss, the Laplacian~\cite{Ilg2018UncertaintyEA} or Gaussian~\cite{Nix1994EstimatingTM} distribution can be adopted for maximising the log-likelihood. More precisely, $\mathbf{\sigma}$ then represents the variance caused by noise inherent in the data. 
Here, we use the information provided by the prediction variance $\sigma$ and define the auxiliary loss function $\mathcal{L}_{aux,\sigma}$ as a combination of $\mathcal{L}_{aux}$ and $\mathbf{\sigma}$: 
\begin{equation}\label{eq:lauxsigma}
    \mathcal{L}_{aux,\sigma} = \mathcal{L}_{aux} + \lambda \mathbf{\sigma}^{2},
\end{equation}
where $\lambda$ is a hyperparameter that weights the influence of the variance. In this case, not only the inconsistency between the depth and the reference depth but also the predicted variance contributes to the extracted gradients.

\subsection{Gradient Extraction}
The gradients can be extracted with the reference depth $\mathbf{d}_{ref}$ and the auxiliary loss function $\mathcal{L}_{aux}$. Given the depth decoder $\psi(\cdot)$ with a set of $L$ layers $\mathcal{A}=\{1,\dots,L\}$, we only assume access to the feature representations $\mathbf{a} = \{ \mathbf{a}_{i}\}_{i=1}^{L}$ of the depth decoder $\psi(\cdot)$, where $i$ denotes the $i$-th of $L$ decoder layers. 
Importantly, we consider an already trained neural network where the model weights are fixed. This makes our method a \textit{post hoc} uncertainty estimation approach. Furthermore, our approach does not include any kind of training. For that reason, it can be applied to different kinds of encoder-decoder depth estimation models regardless of the model training strategy, i.e., whether the model has been trained with ground truth depth $\mathbf{d}$ in a supervised manner or self-supervised with monocular sequences or stereo pairs. Moreover, we do not modify the model architecture and do not need additional modules, which makes our method applicable to various model architectures, i.e., to fully convolutional or transformer-based models. 
The auxiliary loss function $\mathcal{L}_{aux}$ is back-propagated through the depth decoder $\psi(\cdot)$. In contrast to the gradient generation for parameter updates, we compute the derivative of the loss \textit{w.r.t.} the feature representations $\mathbf{a}_{i}$:
\begin{equation}
    \mathbf{g}_{i} = \frac{\partial \mathcal{L}_{aux}}{\partial \mathbf{a}_{i}},
\end{equation}
where $\mathbf{g}_{i} \in \mathbb{R}^{w_{g,i} \times h_{g,i} \times c_i}$ are the resulting gradient maps with width $w_{g,i}$, height $h_{g,i}$ and number of channels $c_i$. Note that $w_{g,i}$, $h_{g,i}$, and $c_i$ are different depending on the decoder layer. The extracted gradient maps are then used to calculate the pixel-wise uncertainty score.
\subsection{Pixel-wise Uncertainty Score}\label{sec:uct_score}
The extracted gradients $\mathbf{g}_{i}$ are used to build the uncertainty score $\mathbf{u}$ either from a single layer $\mathbf{a}_{i}$ with $i \in \mathcal{A}$ or from $k$ layers $\{\mathbf{a}_{i}\}_{i=1}^{k}$, where the $k$ layers are a subset $\mathcal{K} \in \mathcal{A}$ of all decoder layers with $|\mathcal{K}| = k$. 
For both approaches, all gradient maps must be processed first because the uncertainty value $\mathbf{u}$ is defined per pixel and the gradient maps $\mathbf{g}_{i}$ have a larger number of channels and a lower resolution. In addition, the gradient maps do not have a defined value range, which is why we normalise them to the range of $[0,1]$. This is solved using the three following operations. To reduce the number of channels in the gradient map $\mathbf{g}_{i}$ from $c$ to $1$, we use the max-pooling operation
\begin{equation}\label{eq:max}
    max_{c}(\cdot):  \mathbb{R}^{w_{g,i}, \times h_{g,i} \times c_i} \to \mathbb{R}^{w_{g,i} \times h_{g,i} \times 1}
\end{equation}
over the channel dimension. 
The gradient with a high magnitude in the respective channel dimension is preferred because it highlights the largest error in the respective spatial region of the gradient map. 
Then, the gradient map is upsampled with bilinear interpolation to the spatial resolution of the depth map: 
\begin{equation}\label{eq:up}
    upsample(\cdot): \mathbb{R}^{w_{g} \times h_{g} \times 1} \to \mathbb{R}^{w \times h \times 1}.
\end{equation}
Lastly, self-normalisation with the minimum and maximum tensor values is applied to obtain gradients in the range of $[0, 1]$:  
\begin{equation}\label{eq:norm}
    normalize(\mathbf{t})=\frac{\mathbf{t} - \min(\mathbf{t})}{\max(\mathbf{t}) - \min(\mathbf{t})},
\end{equation}
where $\mathbf{t}$ is an arbitrary tensor.  
\subsubsection{Uncertainty Score Using a Single Layer}
When a single layer is chosen for the gradient extraction, Eq.~\ref{eq:max} to Eq.~\ref{eq:norm} are applied to a single gradient map one after another: 
\begin{equation}\label{eq:uctonelayer}
    \mathbf{u}_{i} = normalize(upsample(max_{c}(\mathbf{g}_{i}))),
\end{equation}
where $\mathbf{u} = \mathbf{u}_{i}$ is the final pixel-wise uncertainty score. 
\subsubsection{Uncertainty Score Using Multiple Layers}
\label{sec:gradex}
Since using a single layer requires careful layer selection, we rely on the uncertainty score from $k$ layers as an alternative option. In this case, $k$ from the $L$ decoder layers are utilised. Then, Eq.~\ref{eq:uctonelayer} is applied to the extracted gradients from all selected layers separately to obtain the uncertainty maps $\{\mathbf{u}_{i}\}_{i=1}^{k}$ from the gradient maps $\{\mathbf{g}_{i}\}_{i=1}^{k}$. The resulting uncertainty maps can then be concatenated to $\mathbf{u}_{conc} = [\{\mathbf{u}_{i}\}_{i=1}^{k}]$ with $\mathbf{u}_{conc} \in \mathbb{R}^{k \times w \times h \times 1}$. 
To remove the layer dimension $k$, we use max-pooling over the uncertainty maps
\begin{equation}\label{eq:maxk}
max_{k}(\cdot):\mathbb{R}^{k \times w \times h \times 1} \to \mathbb{R}^{w \times h \times 1}. 
\end{equation}
In this context, Eq.~\ref{eq:maxk} is applied to $\mathbf{u}_{con}$ in order to obtain the final uncertainty map $\mathbf{u}$: 
\begin{equation}
    \mathbf{u} = max_{k}(\mathbf{u}_{conc}).
\end{equation}
Since we choose the $k$ last layers, this removes the layer selection and makes the gradient extraction more robust and independent of a specific decoder layer. 

\subsection{Discussion on Reference Depth}
Next to generating the reference depth in the image space, we examine an alternative option to generate the reference depth using reference features instead of a reference image. The reference features $\mathbf{z}^{\prime}$, which are a transformed version of the encoded features $\mathbf{z}$ serve as input to the depth decoder.
First, the input image $\mathbf{x}$ is passed through the encoder to obtain the features $\mathbf{z}=\phi(\mathbf{x})$. As for the reference image, we define the transformation function $T(\cdot)$ to obtain the reference features $\mathbf{z}^{\prime}=T(\mathbf{z})$. Then, the transformed features are forwarded through the depth decoder to obtain the corresponding depth prediction $\mathbf{d}^{\prime} = \psi(\mathbf{z}^{\prime})$. Again, if $T(\cdot)$ is a invertible geometric transformation, the inverse function $T^{-1}(\cdot)$ must be applied to the depth prediction $\mathbf{d}^{\prime}$ to obtain the final reference depth $\mathbf{d}_{ref}$. Otherwise, the predicted depth $\mathbf{d}^{\prime}$ serves already as the reference depth $\mathbf{d}_{ref}$.
We show results using the reference features instead of the image in our ablation studies in Section~\ref{sec:imgtransform}. \\

\section{Experiments}
We first introduce the uncertainty estimation metrics to evaluate our uncertainty estimation approach on two standard depth estimation benchmarks. Furthermore, we demonstrate the effectiveness of our gradient-based uncertainty estimation in comparison to related work. Afterwards, we assess different design choices of our method through various ablation studies. 

\subsection{Uncertainty Evaluation Metrics}
\textbf{Normalised Uncertainty Calibration Error:}
In~\cite{Laves2019WellcalibratedMU}, the uncertainty calibration error (UCE) was introduced as a modification from the expected calibration error (ECE)~\cite{Naeini_Cooper_Hauskrecht_2015}, which measures the agreement between the confidence and the accuracy for classification models. 
In this paper, we introduce the normalised UCE (nUCE) as a modification by normalising the error and the uncertainty for the UCE calculation. This is motivated by the fact that all uncertainty methods output a different range of uncertainty, e.g., uncertainty is given as a physical unit or unitless in an arbitrary range. We argue that the uncertainty should correspond to the true error, and therefore the pixel with the largest uncertainty should correspond to the largest error, regardless of the output range of the uncertainty. In regression tasks, such as monocular depth estimation, it is important to highlight regions of high error regardless of the exact error unit, since accurate error correction is not feasible. Therefore, we normalise the error $e_{j} = || \hat{d}_{j} - d_{j} ||^{2}$ and the uncertainty $u_{j}$ to the range of $[0, 1]$ before calculating the nUCE: 
\begin{equation}
    e_{j,norm} = \frac{e_{j} - \min(\mathcal{E})}{\max(\mathcal{E}) - \min(\mathcal{E})},  
\end{equation}
\begin{equation}
    u_{j,norm} = \frac{u_{j} - \min(\mathcal{U})}{\max(\mathcal{U}) - \min(\mathcal{U})},  
\end{equation}
with the set of all errors $\mathcal{E}=\{e_{j}\}_{j=1}^{n}$ and the set of all uncertainties $\mathcal{U}=\{u_{j}\}_{j=1}^{n}$. 
Note that the single error and uncertainty values are the error and uncertainty on the pixel level. The error range is then divided into $M$ bins $B_{m}$, where the error per bin $error(B_{m})$ and the uncertainty per bin $uncert(B_{m})$ are then defined as: 
\begin{equation}
    error(B_m)=\frac{1}{|B_{m}|}\sum_{j\in B_{m}}e_{j,norm}, 
\end{equation}
\begin{equation}
    uncert(B_m)=\frac{1}{|B_{m}|}\sum_{j\in B_{m}}u_{j,norm}.
\end{equation}
Finally, the error and uncertainty per bin can be used to obtain the nUCE similar to the UCE but with the important difference that the error and uncertainty values are normalised: 
\begin{equation}
    \text{nUCE} = \sum_{m=1}^{M} \frac{|B_{m}|}{n}|error(B_{m}) - uncert(B_{m})|, 
\end{equation}
where $n$ is the total number of inputs. 
\\
\\
\textbf{Sparsification Plots:}
Following the literature~\cite{Ilg2018UncertaintyEA,Poggi2020OnTU}, we rely on the commonly used sparsification curves to evaluate the estimated uncertainties.
Therefore, we compute the oracle sparsification, the random sparsification, and the actual sparsification. The sparsification curves are calculated for one test image given an error metric and are then averaged over the test set. We compute the sparsification curves in terms of the standard depth estimation metrics absolute relative error (Abs Rel), root mean squared error (RMSE), and accuracy ($\delta \geq 1.25$). To obtain the oracle or actual sparsification curves, the pixels are ordered by descending error and uncertainty, respectively. Then, a specified fraction of pixels with the highest error/uncertainty is removed and the respective error metric is calculated with the remaining pixels. If the uncertainty estimate matches the true error, the error metric improves for the remaining pixels. In contrast, with random sparsification, the average error remains unchanged when fractions of pixels are removed. Finally, the sparsification error is the difference between the oracle sparsification and the actual sparsification. Therefore, it measures whether the estimated uncertainty agrees with the true error. The random gain, on the other hand, is the difference between the actual sparsification and the random sparsification and states whether the uncertainty estimate is better than not modelling the uncertainty. As uncertainty metrics, the area under the sparsification error (AUSE) and the area under the random gain (AURG) are calculated. 
\subsection{Experimental Setup}
\textbf{Datasets:} Similar to~\cite{Poggi2020OnTU}, we evaluate the uncertainty estimation with the KITTI dataset~\cite{Geiger2013IJRR}. KITTI is an autonomous driving dataset consisting of 61 scenes with a maximum depth of 80 meters. The average image resolution is \mbox{$375 \times 1242$}. We use the Eigen split~\cite{Eigen2014DepthMP} with improved ground truth as given in~\cite{Uhrig2017SparsityIC}. Next to KITTI, we propose to evaluate the uncertainty estimation performance on the NYU Depth V2 dataset~\cite{SilbermanECCV12}. This is an indoor dataset recorded in 494 locations with a resolution of $480 \times 640$. The maximum depth is set to 10 meters, which is comparatively small in contrast to KITTI.  \\
\\
\textbf{Models:} For a fair comparison to the different uncertainty estimation approaches in~\cite{Poggi2020OnTU}, we choose Monodepth2~\cite{monodepth2} as the depth estimation model. In addition, we extend the evaluation and implement different uncertainty estimation approaches using the recently introduced transformer-based model MonoViT~\cite{monovit}. We consider both models trained on the KITTI~\cite{Geiger2013IJRR} dataset in a self-supervised manner with monocular sequences as well as with stereo pairs. With monocular sequences, the camera pose must be estimated together with the depth, which introduces additional uncertainties in estimating the pose. Next to the self-supervised settings, we train Monodepth2~\cite{monodepth2} in a supervised manner on NYU Depth V2~\cite{SilbermanECCV12}.\\
\\
\textbf{Comparison to Related Work:} We compare our approach to aleatoric as well as epistemic uncertainty estimation methods.
The considered uncertainty estimation approaches can be divided into trainable methods and \textit{post hoc} methods, such as our approach.
As trainable methods we take Bootstrapped Ensembles (\textit{Boot}~\cite{Lakshminarayanan2017SimpleAS}), Monte Carlo Dropout (\textit{Drop}~\cite{mcdropout}), log-likelihood maximisation (\textit{Log}~\cite{Ilg2018UncertaintyEA,Klodt2018SupervisingTN}) and self-teaching (\textit{Self}~\cite{Poggi2020OnTU}) into account.
While \textit{Drop} and \textit{Boot} predict the epistemic uncertainty, \textit{Log} and \textit{Self} target the aleatoric uncertainty.
Furthermore, \textit{Drop} and \textit{Boot} are based on sampling, whereas \textit{Log} and \textit{Self} predict the uncertainty in a single forward pass. For this reason, \textit{Log} and \textit{Self} can be combined with \textit{post hoc} methods. 
Therefore, in the case of \textit{Log} and \textit{Self}, we differentiate between the model and the uncertainty obtained from the model. While the models are marked with \textit{Log-model} and \textit{Self-model}, the predicted uncertainty obtained from the respective method is highlighted using \textit{Log} and \textit{Self}.
As \textit{post hoc} approaches, we compare our method to post-processing (\textit{Post}~\cite{monodepth2}), BayesCap (\textit{BCap}~\cite{Upadhyay2022BayesCapBI}), inference-only dropout (\textit{Drop}$^\ast$~\cite{Mi2019TrainingFreeUE}) and the variance over different augmentations (\textit{Var}~\cite{Mi2019TrainingFreeUE}).
Since both, \textit{Log} and \textit{Self}, can be combined with \textit{post hoc} methods, we evaluate different combinations of trainable methods and \textit{post hoc} methods.
In addition, we evaluate all \textit{post hoc} methods using a regular depth estimation model~(\textit{Reg-model}) which is trained to only predict depth without uncertainty.
Exceptions are the two \textit{post hoc} methods \textit{Post} and \textit{BCap} which are only applied to the regular depth estimation model~(\textit{Reg-model}).
For \textit{Post}, this is because this method is used as a simple baseline uncertainty estimate. 
For \textit{BCap}, this is since this method also predicts aleatoric uncertainty similarly to \textit{Log} and \textit{Self}.
In summary, our evaluation contains a combination of different depth estimation models and uncertainty estimation methods.
The model can be one from \textit{Reg-model}, \textit{Log-model} and \textit{Self-model}.
The method is one of the considered trainable or \textit{post hoc} uncertainty estimation approaches, namely \textit{Drop}, \textit{Boot}, \textit{Post}, \textit{BCap}, \textit{Drop$^\ast$}, \textit{Var}, \textit{Log}, \textit{Self} or our method (\textit{Ours}). Note that \textit{Drop} and \textit{Boot} provide both depth and uncertainty. \\ 
\\
\textbf{Implementation Details:} All models are trained with Monodepth2~\cite{monodepth2} or MonoViT~\cite{monovit} as depth estimation architecture. For our evaluations, where Monodepth2 is trained on KITTI with monocular or stereo pair supervision, we use the model weights provided by Poggi \textit{et al.}~\cite{Poggi2020OnTU}. All other models are trained from scratch. Note that we use the \textit{Self} method only for the Monodepth2 models trained using KITTI, since in this case the model weights are publicly available. The training of MonoViT follows the same training protocol as for Monodepth2. Moreover, we train Monodepth2 in a supervised manner on NYU Depth V2~\cite{SilbermanECCV12} with the images cropped to a final resolution of $224 \times 288$ and rotation, scaling, horizontal flipping and colour jittering applied randomly beforehand. For \textit{Boot}, we train eight ensembles each. For \textit{Drop}, we train the models with dropout applied after \textit{ConvBlocks} in the decoder. For a fair comparison, we also sample eight times in the case of \textit{Drop} and \textit{Drop}$^\ast$ during inference with a dropout probability of 0.2. The \textit{Log} models are trained with the introduced loss functions from \cite{Ilg2018UncertaintyEA} and \cite{Klodt2018SupervisingTN} for supervised and self-supervised training, respectively. In the case of \textit{BCap}, we follow the provided implementations from~\cite{Upadhyay2022BayesCapBI}. As suggested, in the supervised settings we use the available ground truth information, while we rely on the predicted depth estimate for the self-supervised settings to train the \textit{BCap} model. For \textit{Var}, we use the variance over the predicted depth and the depth predictions for the augmented versions where the original image is augmented by horizontal flipping, grey scaling, rotation, and noise. 

In our approach, we consider horizontal flipping, additive noise, rotation, grey scaling, and diffusion-based augmentation as image space transformations to obtain the reference depth. As feature augmentations, we take horizontal flipping and additive noise into account. By default, we choose the horizontal flip transformation but we investigate the influence of different augmentations in Section~\ref{sec:imgtransform}. We provide our results for three different versions. \textit{Ours} is the gradient-based uncertainty estimation where horizontal flipping is applied to the input image to obtain the reference depth and the uncertainty score is calculated from a single decoder layer. \textit{Ours}$^\dagger$ is similar, but the reference depth is generated with horizontal flipping applied to the features. For Monodepth2, we consider all decoder layers besides the last layer while for MonoViT we only consider decoder layers that are not followed by an attention layer and also exclude the last layer. In both cases, we use the 6-th decoder layer for gradient extraction on Monodepth2, but the 5-th decoder layer in the case of MonoViT. By contrast, for \textit{Ours}$^\ddagger$, the reference depth is obtained with horizontal flipping applied to the input image but the uncertainty score is obtained from multiple layers. Here, for both model architectures, we use the last four layers excluding the decoder layer before the prediction. In the case of the predictive depth estimation models, we set $\lambda$ to $2.0$ in the auxiliary loss function $\mathcal{L}_{aux, \sigma}$. 

\subsection{Uncertainty Estimation Results}\label{sec:exp_results}
\subsubsection{KITTI Monodepth2}
In this section, we provide the uncertainty estimation results in comparison to related work for Monodepth2~\cite{monodepth2} trained in a self-supervised manner with monocular or stereo pair supervision on the KITTI dataset~\cite{Geiger2013IJRR}.\\
\\
\textbf{Monocular Supervision:} In Table~\ref{tab:kitti_mono}, the results obtained for models trained with monocular sequences are provided. For all three models (\textit{Reg-model}, \textit{Log-model} and \textit{Self-model}) all of our gradient-based uncertainty estimation variants obtain better results in all metrics and therefore provide improved uncertainty estimation maps over the uncertainty provided by the model itself or the other \textit{post hoc} approaches. The advantage is especially high for the AUSE and AURG in terms of RMSE and for the nUCE. The worst performance can be observed for \textit{Drop}, \textit{Boot}, or \textit{BCap}, where the AURG is even negative, which means that the uncertainty estimation is worse than not modelling the uncertainty. \textit{BCap} usually relies on the ground truth depth, which is not available for the self-supervised trained models. \\
\\
\textbf{Stereo Pair Supervision:} Table~\ref{tab:kitti_stereo} reports the uncertainty estimation performance for models trained with stereo pair supervision. Here, the uncertainty comes only from the depth estimate and not from the depth and the pose estimation as in monocular training. Again, our gradient-based approach leads to uncertainty estimates that better match the error, which is especially visible in the nUCE, where the distribution of the error is taken into account. While \textit{Self} and our approach are on par in the AUSE and AURG in terms of Abs Rel and $\delta \geq 1.25$, our method is a \textit{post hoc} approach, whereas \textit{Self} requires the training of two models to obtain the final model.  Here, the uncertainty outputs of \textit{Log} and \textit{Boot} are comparably improved to the monocular-trained model. For \textit{BCap}, although the uncertainty corresponds to the true error in certain error ranges, the highest uncertainty does not necessarily correspond to the highest error. This is reflected in the comparatively worse results for all metrics except nUCE, which is weighted by the error distribution. 
\setlength{\tabcolsep}{0.5pt}
\begin{table}[ht]
    \centering
    \caption{Uncertainty evaluation results for Monodepth2~\cite{monodepth2} trained with monocular supervision on KITTI~\cite{Geiger2013IJRR}. The first column specifies the model used for depth estimation, i.e., a regular depth estimation model (\textit{Reg-model}) or predictive depth estimation models that already predict an uncertainty (\textit{Log-model},\textit{Self-model}). The second column contains the different uncertainty estimation methods applied to the respective depth estimation models. Note that \textit{Drop} and \textit{Boot} provide depth and uncertainty, but no \textit{post hoc} approach is applied since the uncertainty is obtained with multiple forward passes.} 
    \begin{tabular}
    {+L{10pt}L{39pt}C{26pt}C{28pt}C{26pt}C{28pt}C{26pt}C{28pt}C{26pt}+}
        \toprule
        & & \multicolumn{2}{c}{Abs Rel} & \multicolumn{2}{c}{RMSE} & \multicolumn{2}{c}{$\delta \geq 1.25$} & \\
        \midrule 
        & & AUSE$^\downarrow$ & AURG$^\uparrow$ &  AUSE$^\downarrow$ & AURG$^\uparrow$ &  AUSE$^\downarrow$ & AURG$^\uparrow$ & nUCE$^\downarrow$ \\
        \midrule 
        & Drop~\cite{Poggi2020OnTU} & 0.066 & 0.000 & 2.614 & 0.981 & 0.098 & 0.003 & 0.149 \\
        & Boot~\cite{Poggi2020OnTU} & 0.058 & 0.000 & 3.991 & -0.747 & 0.084 & -0.002 & 0.036 \\
        \midrule
        \multirow{7}{*}{\rotatebox{90}{Reg-model}} & Post~\cite{monodepth2} & 0.044 & 0.012 & 2.867 & 0.411 & 0.056 & 0.022 & 0.015 \\
        & BCap~\cite{Upadhyay2022BayesCapBI} & 0.068 & -0.010 & 3.717 & -0.351 & 0.098 & -0.018 & 0.440 \\
        & Drop$^\ast$~\cite{Mi2019TrainingFreeUE} & 0.031 & 0.027 & 0.854 & 2.512 & 0.029 & 0.051 & 0.153 \\
        & Var~\cite{Mi2019TrainingFreeUE} & 0.055 & 0.003 & 3.575 & -0.207 & 0.074 & 0.006 & 0.055 \\
        & \cellcolor{grays}Ours & \cellcolor{grays}0.029 & \cellcolor{grays}0.029 & \cellcolor{grays}\textbf{0.533} & \cellcolor{grays}\textbf{2.833} & \cellcolor{grays}\textbf{0.025} & \cellcolor{grays}\textbf{0.055} & \cellcolor{grays}0.005 \\
        & \cellcolor{grays}Ours$^\dagger$ & \cellcolor{grays}\textbf{0.028} & \cellcolor{grays}\textbf{0.030} & \cellcolor{grays}0.542 & \cellcolor{grays}2.824 & \cellcolor{grays}0.026 & \cellcolor{grays}0.054 & \cellcolor{grays}\textbf{0.004} \\
        & \cellcolor{grays}Ours$^\ddagger$ & \cellcolor{grays}0.029 & \cellcolor{grays}0.029 & \cellcolor{grays}0.548 & \cellcolor{grays}2.819 & \cellcolor{grays}0.026 & \cellcolor{grays}0.054 & \cellcolor{grays}0.005 \\ 
        \midrule
         & Log~\cite{Poggi2020OnTU} & 0.039 & 0.020 & 2.563 & 0.916 & 0.044 & 0.038 & 0.317 \\
        & Drop$^\ast$~\cite{Mi2019TrainingFreeUE} & 0.048 & 0.011 & 2.049 & 1.430 & 0.062 & 0.021 & 0.126 \\
        & Var~\cite{Mi2019TrainingFreeUE} & 0.055 & 0.004 & 3.687 & -0.208 & 0.075 & 0.008 & 0.055 \\ 
         & \cellcolor{grays}Ours & \cellcolor{grays}\textbf{0.026} & \cellcolor{grays}\textbf{0.033} & \cellcolor{grays}0.824 & \cellcolor{grays}2.655 & \cellcolor{grays}\textbf{0.024} & \cellcolor{grays}\textbf{0.059} & \cellcolor{grays}0.006 \\
        & \cellcolor{grays}Ours$^\dagger$ & \cellcolor{grays}\textbf{0.026} & \cellcolor{grays}0.032 & \cellcolor{grays}\textbf{0.767} & \cellcolor{grays}\textbf{2.712} & \cellcolor{grays}\textbf{0.024} & \cellcolor{grays}\textbf{0.059} & \cellcolor{grays}\textbf{0.005} \\
        \multirow{-6}{*}{\rotatebox{90}{Log-model}} & \cellcolor{grays}Ours$^\ddagger$ & \cellcolor{grays}\textbf{0.026} & \cellcolor{grays}\textbf{0.033} & \cellcolor{grays}0.813 & \cellcolor{grays}2.666 & \cellcolor{grays}\textbf{0.024} & \cellcolor{grays}\textbf{0.059} & \cellcolor{grays}0.007\\
        \midrule
         & Self~\cite{Poggi2020OnTU} & 0.030 & 0.026 & 2.012 & 1.264 & 0.030 & 0.045 & 0.033 \\
        & Drop$^\ast$~\cite{Mi2019TrainingFreeUE} & 0.034 & 0.022 & 1.145 & 2.131 & 0.033 & 0.041 & 0.160 \\
        & Var~\cite{Mi2019TrainingFreeUE} & 0.055 & 0.001 & 3.631 & -0.355 & 0.073 & 0.002 & 0.056 \\
          & \cellcolor{grays}Ours & \cellcolor{grays}\textbf{0.024} & \cellcolor{grays}\textbf{0.033} & \cellcolor{grays}0.491 & \cellcolor{grays}2.784 & \cellcolor{grays}\textbf{0.017} & \cellcolor{grays}\textbf{0.058} & \cellcolor{grays}0.005 \\
        & \cellcolor{grays}Ours$^\dagger$ & \cellcolor{grays}0.025 & \cellcolor{grays}0.031 & \cellcolor{grays}0.538 & \cellcolor{grays}2.738 & \cellcolor{grays}0.019 & \cellcolor{grays}0.056 & \cellcolor{grays}\textbf{0.004} \\
        \multirow{-6}{*}{\rotatebox{90}{Self-model}} & \cellcolor{grays}Ours$^\ddagger$ & \cellcolor{grays}\textbf{0.024} & \cellcolor{grays}\textbf{0.033} & \cellcolor{grays}\textbf{0.489} & \cellcolor{grays}\textbf{2.787} & \cellcolor{grays}\textbf{0.017} & \cellcolor{grays}\textbf{0.058} & \cellcolor{grays}0.005 \\
        \bottomrule
    \end{tabular}
    \label{tab:kitti_mono}
\end{table}
\setlength{\tabcolsep}{0.5pt}
\begin{table}[ht]
    \centering
    \caption{Uncertainty evaluation results for Monodepth2~\cite{monodepth2} trained with stereo pair supervision on KITTI~\cite{Geiger2013IJRR}. The first column specifies the model used for depth estimation, i.e., a regular depth estimation model (\textit{Reg-model}) or predictive depth estimation models that already predict an uncertainty (\textit{Log-model},\textit{Self-model}). The second column contains the different uncertainty estimation methods applied to the respective depth estimation models. Note that \textit{Drop} and \textit{Boot} provide depth and uncertainty, but no \textit{post hoc} approach is applied since the uncertainty is obtained with multiple forward passes.} 
    \begin{tabular}
    {+L{10pt}L{34pt}C{26pt}C{28pt}C{26pt}C{28pt}C{26pt}C{28pt}C{26pt}+}
    \toprule
        & & \multicolumn{2}{c}{Abs Rel} & \multicolumn{2}{c}{RMSE} & \multicolumn{2}{c}{$\delta \geq 1.25$} & \\
        \midrule
        & & AUSE$^\downarrow$ & AURG$^\uparrow$ &  AUSE$^\downarrow$ & AURG$^\uparrow$ &  AUSE$^\downarrow$ & AURG$^\uparrow$ & nUCE$^\downarrow$\\
        \midrule
        & Drop~\cite{Poggi2020OnTU} & 0.104 & -0.030 & 6.321 & -2.252 & 0.236 & -0.080 & 0.261 \\
        & Boot~\cite{Poggi2020OnTU} & 0.028 & 0.029 & 2.288 & 0.955 & 0.031 & 0.048 & 0.020 \\
        \midrule
        \multirow{7}{*}{\rotatebox{90}{Reg-model}} & Post~\cite{monodepth2} & 0.036 & 0.019 & 2.518 & 0.729 & 0.045  & 0.034 & 0.012 \\
        & BCap~\cite{Upadhyay2022BayesCapBI} & 0.061 & -0.004 & 3.003 & 0.326 & 0.087 & -0.006 & \textbf{0.004} \\
        & Drop$^\ast$~\cite{Mi2019TrainingFreeUE} & 0.061 & -0.004 & 3.492 & -0.163 & 0.089 & -0.009 & 0.148 \\
        & Var~\cite{Mi2019TrainingFreeUE} & 0.053 & 0.004 & 3.640 & -0.310 & 0.073 & 0.007 & 0.052 \\
        & \cellcolor{grays}Ours & \cellcolor{grays}0.022 & \cellcolor{grays}0.035 & \cellcolor{grays}0.513 & \cellcolor{grays}2.816 & \cellcolor{grays}0.023 & \cellcolor{grays}0.057 & \cellcolor{grays}0.005 \\
        & \cellcolor{grays}Ours$^\dagger$ & \cellcolor{grays}0.024 & \cellcolor{grays}0.033 & \cellcolor{grays}0.567 & \cellcolor{grays}2.762 & \cellcolor{grays}0.027 & \cellcolor{grays}0.053 & \cellcolor{grays}0.005 \\
        & \cellcolor{grays}Ours$^\ddagger$ & \cellcolor{grays}\textbf{0.021} & \cellcolor{grays}\textbf{0.036} & \cellcolor{grays}\textbf{0.477} & \cellcolor{grays}\textbf{2.852} & \cellcolor{grays}\textbf{0.021} & \cellcolor{grays}\textbf{0.059} & \cellcolor{grays}0.006 \\
        \midrule
        & Log~\cite{Poggi2020OnTU} & 0.022 & 0.035 & 0.953 & 2.377 & \textbf{0.018} & \textbf{0.061} & 0.282 \\
        & Drop$^\ast$~\cite{Mi2019TrainingFreeUE} & 0.073 & -0.016 & 4.070 & -0.740 & 0.109 & -0.029 & 0.203 \\
        & Var~\cite{Mi2019TrainingFreeUE} & 0.053 & 0.004 & 3.613 & -0.282 & 0.073 & 0.007 & 0.052 \\
        & \cellcolor{grays}Ours & \cellcolor{grays}\textbf{0.019} & \cellcolor{grays}\textbf{0.038} & \cellcolor{grays}\textbf{0.501} & \cellcolor{grays}\textbf{2.830} & \cellcolor{grays}\textbf{0.018} & \cellcolor{grays}\textbf{0.061} & \cellcolor{grays}\textbf{0.005} \\
        & \cellcolor{grays}Ours$^\dagger$ & \cellcolor{grays}0.021 & \cellcolor{grays}0.036 & \cellcolor{grays}0.544 & \cellcolor{grays}2.787 & \cellcolor{grays}0.022 & \cellcolor{grays}0.058 & \cellcolor{grays}\textbf{0.005} \\
        \multirow{-6}{*}{\rotatebox{90}{Log-model}} & \cellcolor{grays}Ours$^\ddagger$ & \cellcolor{grays}\textbf{0.019} & \cellcolor{grays}\textbf{0.038} & \cellcolor{grays}0.510 & \cellcolor{grays}2.820 & \cellcolor{grays}0.019 & \cellcolor{grays}0.061 & \cellcolor{grays}\textbf{0.005} \\ 
        \midrule
        & Self~\cite{Poggi2020OnTU} & 0.022 & 0.035 & 1.668 & 1.629 & \textbf{0.022} & \textbf{0.056} & 0.039 \\
        & Drop$^\ast$~\cite{Mi2019TrainingFreeUE} & 0.075 & -0.019 & 4.123 & -0.827 & 0.110 & -0.032 & 0.230 \\
        & Var~\cite{Mi2019TrainingFreeUE} & 0.055 & 0.002 & 3.677 & -0.380 & 0.075 & 0.003 & 0.053 \\
        & \cellcolor{grays}Ours & \cellcolor{grays}0.022 & \cellcolor{grays}\textbf{0.035} & \cellcolor{grays}0.517 & \cellcolor{grays}2.780 & \cellcolor{grays}0.023 & \cellcolor{grays}0.055 & \cellcolor{grays}0.005 \\
        & \cellcolor{grays}Ours$^\dagger$ & \cellcolor{grays}0.023 & \cellcolor{grays}0.034 & \cellcolor{grays}0.568 & \cellcolor{grays}2.729 & \cellcolor{grays}0.025 & \cellcolor{grays}0.053 & \cellcolor{grays}\textbf{0.004} \\
        \multirow{-6}{*}{\rotatebox{90}{Self-model}} & \cellcolor{grays}Ours$^\ddagger$ & \cellcolor{grays}\textbf{0.021} & \cellcolor{grays}\textbf{0.035} & \cellcolor{grays}\textbf{0.480} & \cellcolor{grays}\textbf{2.817} & \cellcolor{grays}\textbf{0.022} & \cellcolor{grays}\textbf{0.056} & \cellcolor{grays}0.006\\
        \bottomrule 
    \end{tabular}
    \label{tab:kitti_stereo}
\end{table}
\subsubsection{KITTI MonoViT} Next to the fully convolutional Monodepth2 architecture, we also evaluate the uncertainty estimation methods on the recent transformer model MonoViT~\cite{monovit}. \\
\\
\textbf{Monocular Supervision:} 
The uncertainty estimation results for MonoViT trained with monocular supervision are reported in Table~\ref{tab:kitti_mono_monovit}. Similar to Monodepth2, our method shows the best uncertainty estimation results in all metrics. Here, \textit{BCap} shows good values for nUCE, but worse results for the other metrics. Comparing our approach to the results of \textit{Post}, which is similar to using our auxiliary loss function without gradient extraction, demonstrates that the information provided by the gradients is even more useful for uncertainty estimation than using the variance over two outputs alone.  \\
\\
\textbf{Stereo Pair Supervision:} In Table~\ref{tab:kitti_stereo_monovit}, the results for MonoViT trained with stereo pair supervision are outlined. Our gradient-based uncertainty estimation demonstrates good results in all metrics, especially the nUCE. While \textit{Log} achieves the best uncertainty estimation results for AUSE and AURG in terms of Abs Rel and $\delta \geq 1.25$, it does not score well in the case of the nUCE. This means that the uncertainty is in good agreement with the error with respect to the order, but the uncertainty values are not well calibrated. When analysing the \textit{post hoc} methods alone, our gradient-based uncertainty provides significantly improved uncertainty estimates not only in AUSE and AURG but also for nUCE. 
\setlength{\tabcolsep}{0.5pt}
\begin{table}[ht]
    \centering
    \caption{Uncertainty evaluation results for MonoViT~\cite{monovit} trained with monocular supervision on KITTI~\cite{Geiger2013IJRR}. The first column specifies the model used for depth estimation, i.e., a regular depth estimation model (\textit{Reg-model}) or a predictive depth estimation model that already predicts an uncertainty (\textit{Log-model}). The second column contains the different uncertainty estimation methods applied to the respective depth estimation models. Note that \textit{Drop} and \textit{Boot} provide depth and uncertainty, but no \textit{post hoc} approach is applied since the uncertainty is obtained with multiple forward passes.}
    \begin{tabular}
    {+L{10pt}L{34pt}C{26pt}C{28pt}C{26pt}C{28pt}C{26pt}C{28pt}C{26pt}+}
        \toprule
        & & \multicolumn{2}{c}{Abs Rel} & \multicolumn{2}{c}{RMSE} & \multicolumn{2}{c}{$\delta \geq 1.25$} & \\
        \midrule 
        & & AUSE$^\downarrow$ & AURG$^\uparrow$ &  AUSE$^\downarrow$ & AURG$^\uparrow$ &  AUSE$^\downarrow$ & AURG$^\uparrow$ & nUCE$^\downarrow$  \\
        \midrule
        & Drop~\cite{mcdropout} & 0.038 & 0.012 & 2.453 & 0.533 & 0.044 & 0.019 & 0.008 \\
        & Boot~\cite{Lakshminarayanan2017SimpleAS} & 0.058 & -0.004 & 4.039 & -0.947 & 0.081 & -0.009 & 0.056 \\
        \midrule
        \multirow{7}{*}{\rotatebox{90}{Reg-model}} & Post~\cite{monodepth2} & 0.041 & 0.008 & 2.684 & 0.260 & 0.046 & 0.014 & 0.016 \\
        & BCap~\cite{Upadhyay2022BayesCapBI} & 0.059 & -0.009 & 3.469 & -0.507 & 0.083 & -0.022 & \textbf{0.004} \\
        & Drop$^\ast$~\cite{Mi2019TrainingFreeUE} & 0.034 & 0.016 & 1.670 & 1.292 & 0.033 & 0.028 & 0.043 \\
        & Var~\cite{Mi2019TrainingFreeUE} & 0.056 & -0.006 & 3.750 & -0.788 & 0.073 & -0.011 & 0.045 \\
        & \cellcolor{grays}Ours & \cellcolor{grays}0.028 & \cellcolor{grays}0.022 & \cellcolor{grays}0.657 & \cellcolor{grays}2.305 & \cellcolor{grays}0.022 & \cellcolor{grays}0.039 & \cellcolor{grays}\textbf{0.004} \\
        & \cellcolor{grays}Ours$^\dagger$ & \cellcolor{grays}\textbf{0.026} & \cellcolor{grays}\textbf{0.024} & \cellcolor{grays}\textbf{0.512} & \cellcolor{grays}\textbf{2.450} & \cellcolor{grays}\textbf{0.019} & \cellcolor{grays}\textbf{0.043} & \cellcolor{grays}\textbf{0.004} \\
         & \cellcolor{grays}Ours$^\ddagger$ & \cellcolor{grays}0.027 & \cellcolor{grays}0.023 & \cellcolor{grays}0.549 & \cellcolor{grays}2.413 & \cellcolor{grays}\textbf{0.019} & \cellcolor{grays}0.042 & \cellcolor{grays}0.005 \\
        \midrule
        & Log~\cite{Klodt2018SupervisingTN} & 0.036 & 0.015 & 2.261 & 0.759 & 0.036 & 0.027 & 0.332 \\
        & Drop$^\ast$~\cite{Mi2019TrainingFreeUE} & 0.076 & -0.024 & 4.592 & -1.573 & 0.110 & -0.047 & 0.140 \\
        & Var~\cite{Mi2019TrainingFreeUE} & 0.058 & -0.006 & 3.819 & -0.800 & 0.074 & -0.011 & 0.046 \\ 
        & \cellcolor{grays}Ours & \cellcolor{grays}0.028 & \cellcolor{grays}0.024 & \cellcolor{grays}0.870 & \cellcolor{grays}2.149 & \cellcolor{grays}0.020 & \cellcolor{grays}0.043 & \cellcolor{grays}\textbf{0.004} \\
        & \cellcolor{grays}Ours$^\dagger$ & \cellcolor{grays}0.027 & \cellcolor{grays}0.024 & \cellcolor{grays}0.897 & \cellcolor{grays}2.122 & \cellcolor{grays}0.019 & \cellcolor{grays}0.044 & \cellcolor{grays}\textbf{0.004} \\
        \multirow{-6}{*}{\rotatebox{90}{Log-model}} & \cellcolor{grays}Ours$^\ddagger$ & \cellcolor{grays}\textbf{0.024} & \cellcolor{grays}\textbf{0.027} & \cellcolor{grays}\textbf{0.623} & \cellcolor{grays}\textbf{2.396} & \cellcolor{grays}\textbf{0.016} & \cellcolor{grays}\textbf{0.047} & \cellcolor{grays}0.006 \\
        \bottomrule 
    \end{tabular}
    \label{tab:kitti_mono_monovit}
\end{table}
\setlength{\tabcolsep}{0.5pt}
\begin{table}[ht]
    \centering
    \caption{Uncertainty evaluation results for MonoViT~\cite{monovit} trained with stereo pair supervision on KITTI~\cite{Geiger2013IJRR}. The first column specifies the model used for depth estimation, i.e., a regular depth estimation model (\textit{Reg-model}) or a predictive depth estimation model that already predicts an uncertainty (\textit{Log-model}). The second column contains the different uncertainty estimation methods applied to the respective depth estimation models. Note that \textit{Drop} and \textit{Boot} provide depth and uncertainty, but no \textit{post hoc} approach is applied since the uncertainty is obtained with multiple forward passes.}
    \begin{tabular}
    {+L{10pt}L{34pt}C{26pt}C{28pt}C{26pt}C{28pt}C{26pt}C{28pt}C{26pt}+}
        \toprule
        & & \multicolumn{2}{c}{Abs Rel} & \multicolumn{2}{c}{RMSE} & \multicolumn{2}{c}{$\delta \geq 1.25$} & \\
        \midrule 
        & & AUSE$^\downarrow$ & AURG$^\uparrow$ &  AUSE$^\downarrow$ & AURG$^\uparrow$ &  AUSE$^\downarrow$ & AURG$^\uparrow$ & nUCE$^\downarrow$\\
        \midrule 
        & Drop~\cite{mcdropout} & 0.031 & 0.020 & 2.043 & 0.948 & 0.030 & 0.030 & 0.007 \\
        & Boot~\cite{Lakshminarayanan2017SimpleAS} & 0.042 & 0.015 & 3.480 & -0.273 & 0.050 & 0.024 & 0.032 \\
        \midrule
        \multirow{7}{*}{\rotatebox{90}{Reg-model}} & Post~\cite{monodepth2} & 0.036 & 0.015 & 2.449 & 0.496 & 0.036 & 0.024 & 0.010 \\
        & BCap~\cite{Upadhyay2022BayesCapBI} & 0.057 & -0.006 & 2.921 & 0.052 & 0.069 & -0.008 & 0.321 \\
        & Drop$^\ast$~\cite{Mi2019TrainingFreeUE} & 0.081 & -0.029 & 4.768 & -1.796 & 0.106 & -0.044 & 0.250 \\
        & Var~\cite{Mi2019TrainingFreeUE} & 0.055 & -0.003 & 3.774 & -0.802 & 0.068 & -0.006 & 0.045 \\
        & \cellcolor{grays}Ours & \cellcolor{grays}0.028 & \cellcolor{grays}0.023 & \cellcolor{grays}0.665 & \cellcolor{grays}2.307 & \cellcolor{grays}0.023 & \cellcolor{grays}0.038 & \cellcolor{grays}\textbf{0.005} \\
        & \cellcolor{grays}Ours$^\dagger$ & \cellcolor{grays}0.026 & \cellcolor{grays}0.025 & \cellcolor{grays}0.607 & \cellcolor{grays}2.365 & \cellcolor{grays}0.023 & \cellcolor{grays}0.039 & \cellcolor{grays}\textbf{0.005} \\
        & \cellcolor{grays}Ours$^\ddagger$ & \cellcolor{grays}\textbf{0.025} & \cellcolor{grays}\textbf{0.027} & \cellcolor{grays}\textbf{0.561} & \cellcolor{grays}\textbf{2.411} & \cellcolor{grays}\textbf{0.020} & \cellcolor{grays}\textbf{0.041} & \cellcolor{grays}0.007 \\
        \midrule
        & Log~\cite{Klodt2018SupervisingTN} & \textbf{0.022} & \textbf{0.029} & 0.988 & 2.057 & \textbf{0.015} & \textbf{0.048} & 0.314 \\
        & Drop$^\ast$~\cite{Mi2019TrainingFreeUE} & 0.083 & -0.032 & 4.853 & -1.808 & 0.111 & -0.048 & 0.175 \\
        & Var~\cite{Mi2019TrainingFreeUE} & 0.055 & -0.004 & 3.859 & -0.814 & 0.069 & -0.006 & 0.045 \\
        & \cellcolor{grays}Ours & \cellcolor{grays}0.033 & \cellcolor{grays}0.018 & \cellcolor{grays}0.940 & \cellcolor{grays}2.104 & \cellcolor{grays}0.033 & \cellcolor{grays}0.030 & \cellcolor{grays}0.003 \\
        & \cellcolor{grays}Ours$^\dagger$ & \cellcolor{grays}0.032 & \cellcolor{grays}0.019 & \cellcolor{grays}0.827 & \cellcolor{grays}2.218 & \cellcolor{grays}0.031 & \cellcolor{grays}0.032 & \cellcolor{grays}\textbf{0.002} \\
        \multirow{-6}{*}{\rotatebox{90}{Log-model}} & \cellcolor{grays}Ours$^\ddagger$ & \cellcolor{grays}0.026 & \cellcolor{grays}0.025 & \cellcolor{grays}\textbf{0.612} & \cellcolor{grays}\textbf{2.433} & \cellcolor{grays}0.022 & \cellcolor{grays}0.041 & \cellcolor{grays}0.006 \\
        \bottomrule
    \end{tabular}
    \label{tab:kitti_stereo_monovit}
\end{table}
\subsubsection{NYU Depth V2}
Table~\ref{tab:nyu} lists the uncertainty estimation performance for Monodepth2 trained supervised on NYU Depth V2. In the supervised case, there are fewer uncertainties due to the training procedure, since ground truth depth maps are available. 
This also helps the trainable methods \textit{Log} and \textit{BCap}, which in this case learn to derive uncertainty using the available ground truth. Comparing the \textit{post hoc} methods, \textit{Post} delivers stable outcomes, while our gradient-based methods are also on par when observing the AUSE and AURG in terms of Abs Rel and $\delta \geq 1.25$. Comparing all versions of our approach, the feature-based augmentation results in the best nUCE. Furthermore, our gradient-based uncertainty estimation method achieves stable results in all situations, regardless of whether ground truth depth is available. \textit{Drop}$^{\ast}$ performs worst among the \textit{post hoc} approaches and also requires computationally expensive sampling during inference. 
\setlength{\tabcolsep}{0.5pt}
\begin{table}[ht]
    \centering
    \caption{Uncertainty evaluation results for Monodepth2~\cite{monodepth2} trained supervised on NYU Depth V2~\cite{SilbermanECCV12}. The first column specifies the model used for depth estimation, i.e., a regular depth estimation model (\textit{Reg-model}) or a predictive depth estimation model that already predicts an uncertainty (\textit{Log-model}). The second column contains the different uncertainty estimation methods applied to the respective depth estimation models. Note that \textit{Drop} and \textit{Boot} provide depth and uncertainty, but no \textit{post hoc} approach is applied since the uncertainty is obtained with multiple forward passes.}
    \begin{tabular}
    {+L{10pt}L{34pt}C{26pt}C{28pt}C{26pt}C{28pt}C{26pt}C{28pt}C{26pt}+}
        \toprule 
        & & \multicolumn{2}{c}{Abs Rel} & \multicolumn{2}{c}{RMSE} & \multicolumn{2}{c}{$\delta \geq 1.25$} & \\
        \midrule 
        & & AUSE$^\downarrow$ & AURG$^\uparrow$ &  AUSE$^\downarrow$ & AURG$^\uparrow$ &  AUSE$^\downarrow$ & AURG$^\uparrow$ & nUCE$^\downarrow$\\
        \midrule 
        & Drop~\cite{mcdropout} & 0.086 & 0.002 & 0.214 & 0.160 & 0.169 & -0.001 & 0.219 \\
        & Boot~\cite{Lakshminarayanan2017SimpleAS} & 0.057 & 0.026 & 0.173 & 0.168 & 0.095 & 0.058 & 0.124 \\
        \midrule
        \multirow{7}{*}{\rotatebox{90}{Reg-model}} & Post~\cite{monodepth2} & 0.067 & 0.017 & 0.268 & 0.094 & 0.117 & 0.036 & 0.016 \\
        & BCap~\cite{Upadhyay2022BayesCapBI} & 0.068 & 0.018 & \textbf{0.199} & \textbf{0.172} & 0.121 & 0.034 & \textbf{0.005} \\ 
        & Drop$^\ast$~\cite{Mi2019TrainingFreeUE} & 0.086 & 0.000 & 0.217 & 0.154 & 0.166 & -0.012 & 0.223 \\
        & Var~\cite{Mi2019TrainingFreeUE} & 0.063 & 0.023 & 0.225 & 0.146 & 0.112 & 0.043 & 0.043 \\
        & \cellcolor{grays}Ours & \cellcolor{grays}\textbf{0.062} & \cellcolor{grays}\textbf{0.025} & \cellcolor{grays}0.252 & \cellcolor{grays}0.119 & \cellcolor{grays}0.107 & \cellcolor{grays}0.048 & \cellcolor{grays}0.092 \\
        & \cellcolor{grays}Ours$^\dagger$ & \cellcolor{grays}0.064 & \cellcolor{grays}0.022 & \cellcolor{grays}0.274 & \cellcolor{grays}0.098 & \cellcolor{grays}\textbf{0.106} & \cellcolor{grays}\textbf{0.049} & \cellcolor{grays}0.080 \\
        & \cellcolor{grays}Ours$^\ddagger$ & \cellcolor{grays}\textbf{0.062} & \cellcolor{grays}0.024 & \cellcolor{grays}0.256 & \cellcolor{grays}0.115 & \cellcolor{grays}0.107 & \cellcolor{grays}0.048 & \cellcolor{grays}0.122 \\
        \midrule
        & Log~\cite{Ilg2018UncertaintyEA} & 0.056 & 0.029 & \textbf{0.160} & \textbf{0.208} & 0.090 & 0.064 & 0.134 \\
        & Drop$^\ast$~\cite{Mi2019TrainingFreeUE} & 0.087 & -0.002 & 0.213 & 0.155 & 0.166 & -0.012 & 0.260 \\
        & Var~\cite{Mi2019TrainingFreeUE} & 0.062 & 0.023 & 0.221 & 0.146 & 0.107 & 0.047 & \textbf{0.045} \\
        & \cellcolor{grays}Ours & \cellcolor{grays}0.054 & \cellcolor{grays}0.031 & \cellcolor{grays}0.177 & \cellcolor{grays}0.190 & \cellcolor{grays}0.088 & \cellcolor{grays}\textbf{0.067} & \cellcolor{grays}0.091 \\
        & \cellcolor{grays}Ours$^\dagger$ & \cellcolor{grays}0.055 & \cellcolor{grays}0.030 & \cellcolor{grays}0.175 & \cellcolor{grays}0.193 & \cellcolor{grays}0.088 & \cellcolor{grays}\textbf{0.067} & \cellcolor{grays}0.086 \\
        \multirow{-6}{*}{\rotatebox{90}{Log-model}} & \cellcolor{grays}Ours$^\ddagger$ & \cellcolor{grays}\textbf{0.053} & \cellcolor{grays}\textbf{0.032} & \cellcolor{grays}0.176 & \cellcolor{grays}0.191 & \cellcolor{grays}\textbf{0.087} & \cellcolor{grays}\textbf{0.067} & \cellcolor{grays}0.095 \\ 
        \bottomrule 
    \end{tabular}
    \label{tab:nyu}
\end{table}
\subsubsection{Sparsification Error Plots}
\begin{figure*}[!ht]
    \begin{tabular}{ccc}
    \subfloat[KITTI MD Mono\label{fig:monodepth-m}]{\centering\includegraphics[height=121pt]{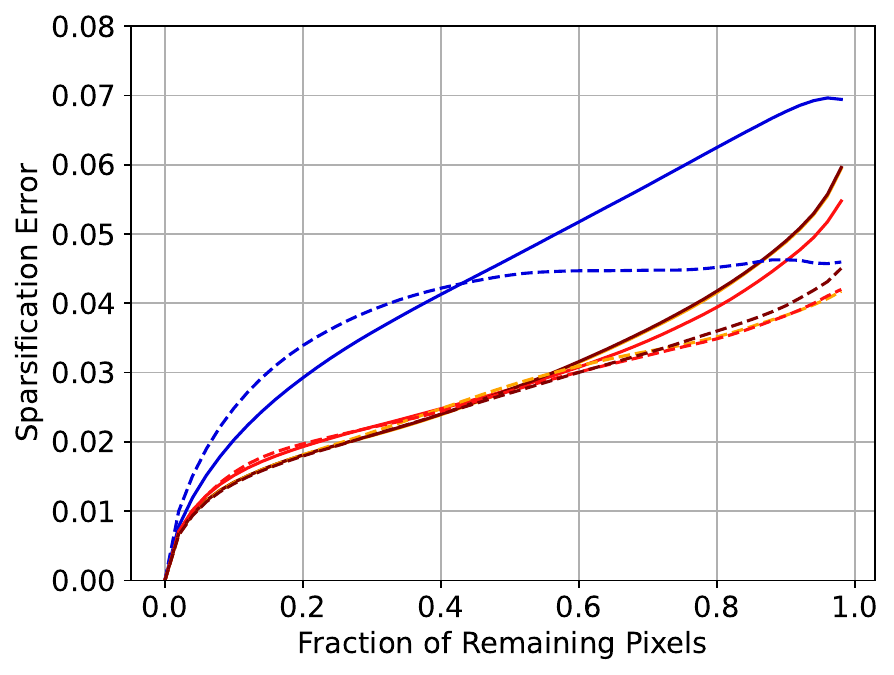}}
    &
    \subfloat[KITTI MD Stereo\label{fig:monodepth-s}]{\centering\includegraphics[height=121pt]{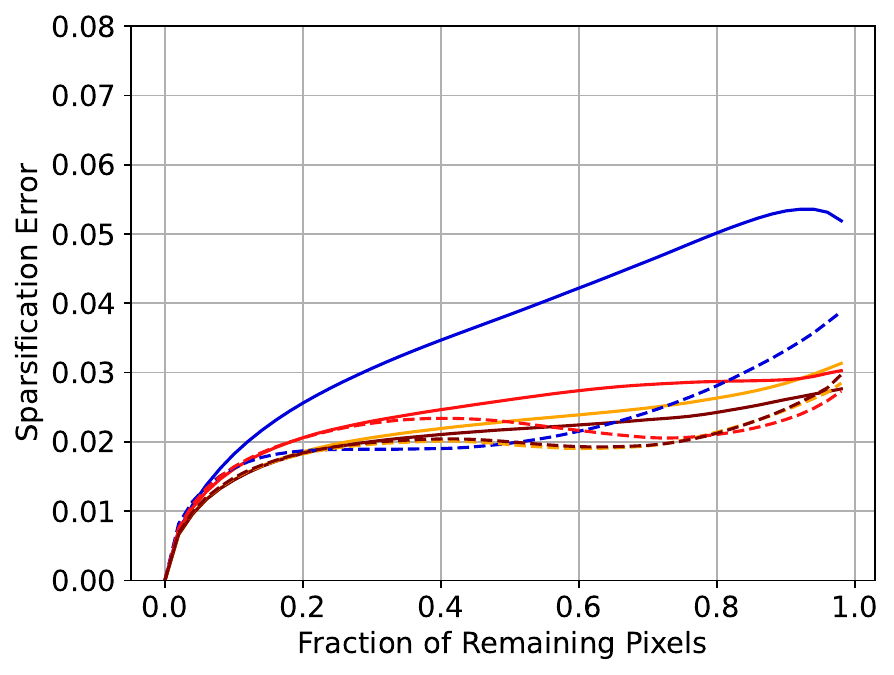}}
    & 
    \subfloat[NYU MD\label{fig:nyu}]{\centering\includegraphics[height=121pt]{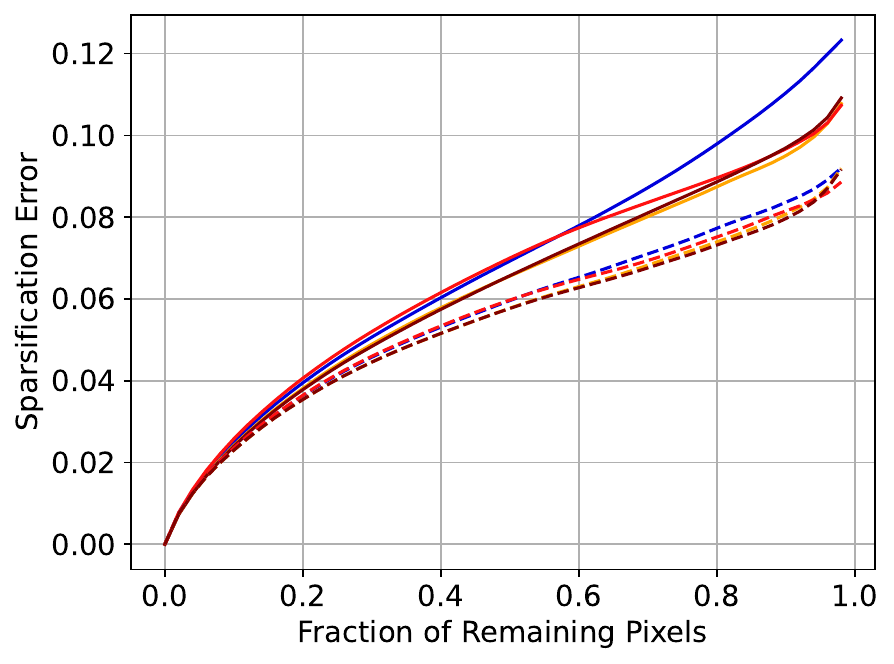}}\\
    \subfloat[KITTI MV Mono\label{fig:monovit-m}]{\centering\includegraphics[height=121pt]{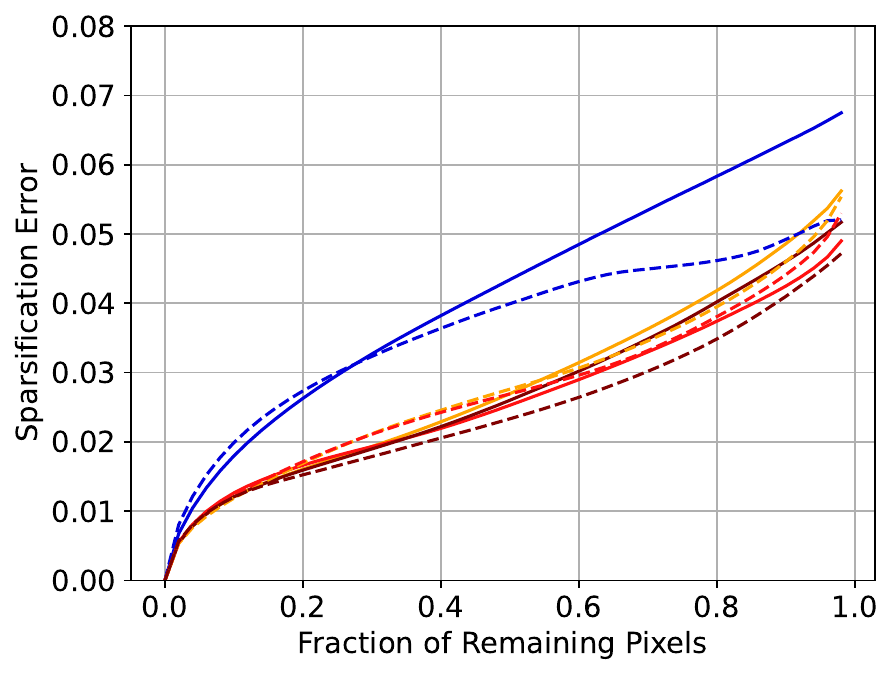}}
    & 
    \subfloat[KITTI MV Stereo\label{fig:monovit-s}]{\centering\includegraphics[height=121pt]{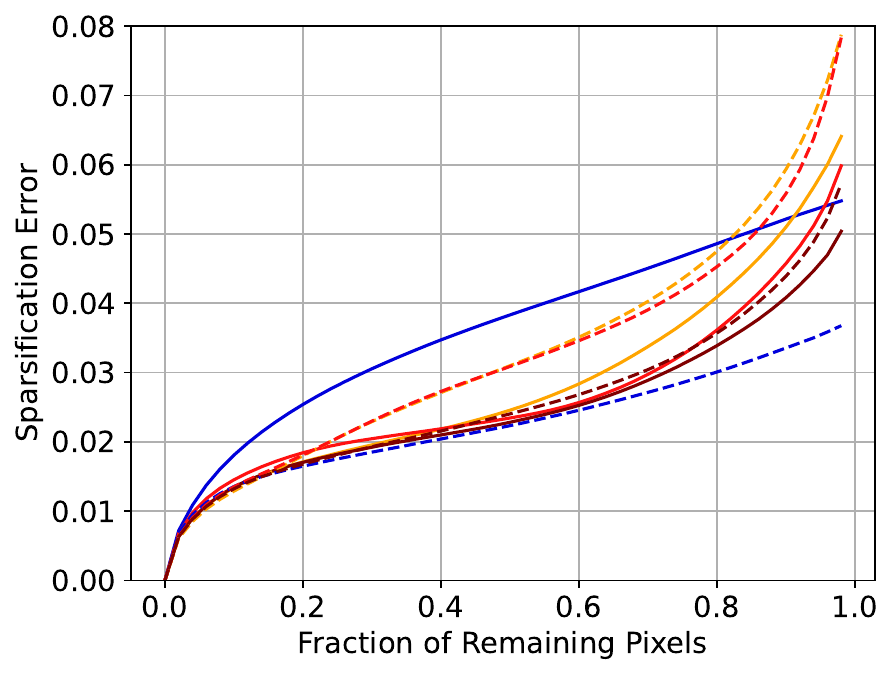}}
    &
    \subfloat{\centering\raisebox{10pt}{\includegraphics[height=110pt]{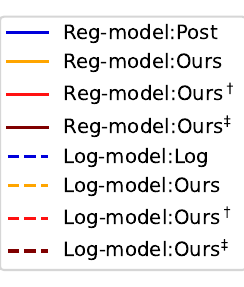}}} \\
    \end{tabular}
    \caption{The sparsification error in terms of Absolute Relative Error (Abs Rel) over the fraction of remaining pixels is shown for Monodepth2~\cite{monodepth2} (MD) and MonoViT~\cite{monovit} (MV) trained with monocular supervision or stereo pair supervision on KITTI~\cite{Geiger2013IJRR} as well as Monodepth2~\cite{monodepth2} trained in a supervised manner on NYU. The model and methods are denoted as [model]:[method]. We compare our gradient-based uncertainty estimation approach to \textit{Post} and \textit{Log} applied to the regular depth estimation model (\textit{Reg-model}) and the predictive depth estimation model (\textit{Log-model}), respectively.}
\label{fig:spars_plots}
\end{figure*}

In Fig.~\ref{fig:spars_plots}, the sparsification errors in terms of the Abs Rel metric are plotted for the regular depth estimation model (\textit{Reg-model}) and the predictive depth estimation model (\textit{Log-model}). This is the deviation of the actual sparsification from the oracle sparsification. We visualise \textit{Post}, \textit{Log} and all versions of our method (Ours). Note that in the diagrams we only consider the methods with the best results in uncertainty estimation. The curves of our method are coloured red and orange, while the other two approaches are marked in blue. A smaller area under the sparsification error curve means that the uncertainty is consistent with the true error for the selected metric. Our gradient-based uncertainty estimates produce curves with the smallest area under the curve, especially in the monocular-trained settings. While the \textit{Log} method shows good performance for NYU and the KITTI models trained with stereo pair supervision, the performance is worse for the KITTI models trained with monocular sequences. \textit{Post}, on the other hand, mostly has a high sparsification error in all graphs. In general, the sparsification plots show that the difference between the different approaches is larger for the self-supervised trained models.

\subsubsection{Visual Results}
\begin{figure*}[ht]
\begin{tabular}{cc@{\hskip1pt}c@{\hskip1pt}cc@{\hskip1pt}c@{\hskip0pt}}
    \subfloat[Image\label{fig:nyu_post_rgb}]{
    \includegraphics[width=0.225\textwidth]{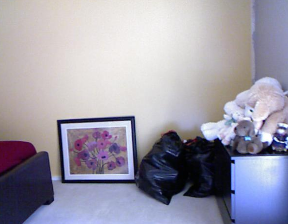}} & 
    \subfloat[Depth\label{fig:nyu_post_depth}]{
    \includegraphics[width=0.225\textwidth]{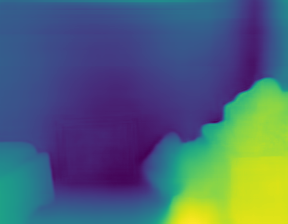}} &
    \hspace{2pt}\includegraphics[height=90.64pt]{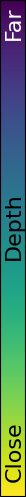} & 
    \subfloat[Error\label{fig:nyu_post_error}]{
    \includegraphics[width=0.225\textwidth]{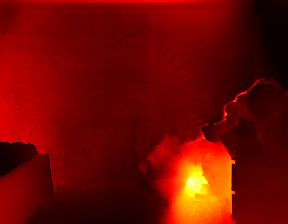}} &
    \subfloat[Ours\label{fig:nyu_post_grad}]{
    \includegraphics[width=0.225\textwidth]{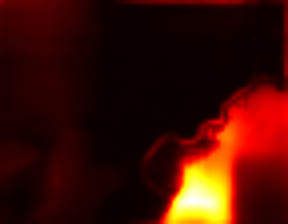}} &
    \hspace{2pt}\includegraphics[height=90.64pt]{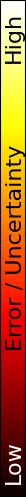} \\
     \subfloat[Post\label{fig:nyu_post_post}]{
    \includegraphics[width=0.225\textwidth]{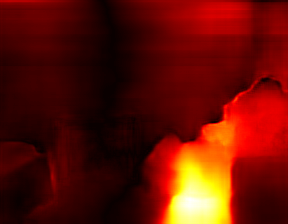}} &
    \subfloat[BCap\label{fig:nyu_post_bcap}]{
    \includegraphics[width=0.225\textwidth]{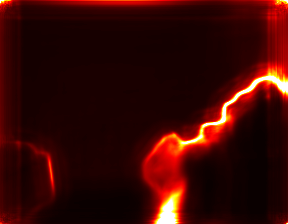}} & &
    \subfloat[Drop$^{\ast}$\label{fig:nyu_post_inferdrop}]{
    \includegraphics[width=0.225\textwidth]{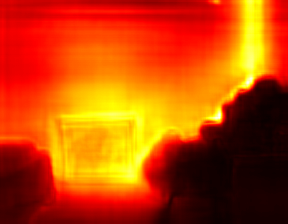}} & 
    \subfloat[Ours$^{\ddagger}$\label{fig:nyu_post_gradm}]{
    \includegraphics[width=0.225\textwidth]{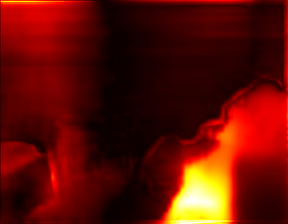}} & 
    \hspace{2pt}\includegraphics[height=90.64pt]{Images/colorbar_hot_error_uct_rot.png} \\
\end{tabular}
    \caption{Uncertainty estimation example from Monodepth2~\cite{monodepth2} trained on NYU Depth V2~\cite{SilbermanECCV12}. In (a), the input image is shown. (b) and (c) visualise the depth prediction and the error, respectively. (d) to (h) display the uncertainty estimates of the different methods.}
    \label{fig:nyu_post_example}
\end{figure*}

Fig.~\ref{fig:nyu_post_example} visualises an example from the NYU Depth V2 dataset with a depth prediction from the regular depth estimation model (\textit{Reg-model}), the squared error, and uncertainty estimates from different approaches. The error map (Fig.~\ref{fig:nyu_post_error}) shows a high error where the garbage bags are located. While \textit{Post} (Fig.~\ref{fig:nyu_post_post}) and our method (Fig.~\ref{fig:nyu_post_grad}, Fig.~\ref{fig:nyu_post_gradm}) estimate this region as most uncertain, \textit{Drop}$^\ast$ (Fig.~\ref{fig:nyu_post_inferdrop}) predicts this region as certain but detects the location of the painting as highest uncertainty. \textit{BCap} (Fig.~\ref{fig:nyu_post_bcap}), on the other hand, mainly detects the edges of the objects on the right side as the most uncertain.  
\setlength{\tabcolsep}{1pt}
\begin{table}[ht]
        \caption{Comparison of the different uncertainty estimation methods regarding the number of trained models (\#Train), whether it is a \textit{post hoc} approach, number of models for uncertainty estimation (\#Models), number of forward passed required to obtain depth and uncertainty (\#FW), number of backward passes (\#BW) and inference time (Inf) in milliseconds [ms] for KITTI Monodepth2 (K-MD) / KITTI MonoViT (K-MV) / NYU (N-MD). Note that $N$ is set to $8$.}
    \centering
    \begin{tabular}{lcccccccccc}
        \toprule
        \multirow{2}{*}{Method} & \multirow{2}{*}{\#Train} & \multirow{2}{*}{PH} & \multirow{2}{*}{\#Models} & \multirow{2}{*}{\#FW} & \multirow{2}{*}{\#BW} & \multicolumn{5}{c}{Inf [ms]}\\
         & & & & & & K-MD & & K-MV & & N-MD \\
        \midrule 
        Log \cite{Poggi2020OnTU} & 1 & No & 1 & 1 & - & 5.89 & / & 29.27 & / & 3.29 \\
        Self \cite{Poggi2020OnTU} & 2 & No & 1 & 1 & - & 5.89 & / & - & / & - \\ 
        \midrule 
        Post \cite{monodepth2} & - & Yes & 1 & 2 & - & 11.78 & / & 58.54 & / & 6.58 \\
        BCap~\cite{Upadhyay2022BayesCapBI} & 1 & Yes & 2 & 1 & - & 9.05 & / & 34.73 & / & 9.12 \\
        Var & - & Yes & 1 & 4 + 1 & - & 29.45 & / & 146.35 & / & 16.45 \\
        Drop$^\ast$ \cite{Mi2019TrainingFreeUE} & - & Yes & 1 & N + 1 & - & 53.01 & / & 263.43  & / & 29.61 \\
        Ours, Ours$^\dagger$ & - & Yes & 1 &  2 & 1 & 18.84 & / & 109.17 & / & 12.05 \\
        Ours$^\ddagger$ & - & Yes & 1 & 2 & 1 & 27.37 & / & 115.75 & / & 18.56 \\ 
        \midrule
        Drop \cite{mcdropout} & 1 & No & 1 & N & - & 47.12 & / & 234.16 & / & 26.32 \\
        Boot \cite{Lakshminarayanan2017SimpleAS} & N & No & N & N & - & 47.12 & / & 234.16 & / & 26.32 \\
        \bottomrule
    \end{tabular}
    \label{tab:method_cmp}
\end{table}
\subsubsection{Discussion}
Table~\ref{tab:method_cmp} lists the key characteristics of the different methods. In particular, we consider the number of models that must be trained (\#Train), whether the method is a \textit{post hoc} approach (PH), the number of models necessary during inference including the depth estimation model (\#Models), the required forward passes (\#FW), the required backward passes (\#BW) and the inference time in milliseconds [ms]. In the case of the inference time, we provide runtimes for Monodepth2 and MonoViT trained on KITTI as well as Monodepth2 trained on NYU. In summary, our method provides stable uncertainty estimation results for all models in all five settings. 
In particular, for the models trained self-supervised with monocular sequences our gradient-based uncertainty shows superior performance. Here the sources of uncertainty are both the depth estimation and the pose estimation.
In safety-critical applications, it is of utmost importance to reduce the computational cost. However, the reliability of the network predictions should be ensured. Thus, the best possible compromise between computational effort and reliability must be found. Although our method adds a slight overhead with the additional backward pass in comparison to the methods that only require one forward pass such as \textit{Log} or \textit{Self}, it is a training-free \textit{post hoc} approach that can be applied to different depth estimation models without retraining or modification of the model. Inference time is especially critical when using a computational more expensive model such as the transformer-based MonoViT. This is clear when comparing the inference times of Monodepth2 and MonoViT trained on KITTI, where the inference time is almost $6\times$ higher. In comparison to the other \textit{post hoc} methods, the additional runtime is still smaller compared to sampling multiple times in the case of \textit{Drop}$^\ast$ or \textit{Var}. Moreover, \textit{Boot} not only has additional overhead during inference but also requires more memory and computational cost due to training $N$ individual models. \textit{BCap}, on the other hand, can provide good results but only with ground truth depth available and requires the training of a second model. 

\subsection{Ablation Studies}
In this section, we assess design choices of methods regarding the selected augmentation to obtain the reference depth, the gradient extraction loss used for the predictive depth estimation models, the layer used for gradient extraction, and the scoring function used to combine gradients extracted from multiple layers. For our ablation studies, we use Monodepth2 and MonoViT trained with monocular sequences on KITTT as well as Monodepth2 trained on NYU. In all ablation studies besides the auxiliary loss used for gradient extraction from the predictive depth estimation model, we conduct the evaluation on the regular depth estimation model. 

\subsubsection{Augmentation for Reference Depth Generation}\label{sec:imgtransform}
In Table~\ref{tab:ref_depth}, we evaluate different augmentations applied to the image and the feature representations to generate the reference depth. As image augmentations, we consider horizontal flipping (Flip), grey-scaling (Gray), additive noise (Noise), rotation with different angles (Rot-[angle]), and diffusion-based augmentation (Diff-R/Diff-S). Importantly, the selected augmentations are not necessarily within the training-time augmentation. 

For diffusion-based augmentation, we rely on a diffusion model trained for domain adaptation. Therefore, we only apply the diffusion-based augmentation to KITTI, where we rely on the corresponding simulated dataset Virtual KITTI~\cite{Gaidon2016Virtual} for training the diffusion model for the domain transfer from KITTI to Virtual KITTI. The domain adaptation is realised by conditioning a Denoising Diffusion Probabilistic Model (DDPM)~\cite{ho2020DenoisingDiffusionProbabilistic} on two domains from the Virtual KITTI dataset~\cite{Gaidon2016Virtual} in a similar fashion to Saharia et al.\cite{saharia2022PaletteImagetoImageDiffusion,hoCascadedDiffusionModels, sahariaPhotorealisticTexttoImageDiffusion2022}. Similar to the colourisation proposed in~\cite{saharia2022PaletteImagetoImageDiffusion}, where the diffusion model is conditioned on colour spaces, we use images from different Virtual KITTI domains instead. In particular, we use the domains \textit{rainy} and \textit{sunset} for the condition of the diffusion model, which is referred to as Diff-R and Diff-S, respectively. Finally, the KITTI test data is augmented using the diffusion model and the resulting domain-transferred KITTI samples are used to generate the reference depth of our uncertainty estimation approach. 

Next to the image space augmentation, we consider horizontal flipping (Flip$^\star$) and additive noise (Noise$^\star$) in the feature space. Since the NYU dataset has ground truth depth available, we also consider the ground truth depth as a reference (GT). For Monodepth2 trained on NYU, ground truth naturally gives the best match to error, except for nUCE. This is because the error metric of nUCE is the squared difference, not the absolute difference. However, in real-world applications, there is no access to any ground truth information. Results across all three settings show that horizontal flip augmentation on an image or feature space is among the best performers in uncertainty estimation. The flipping operation shows the scene or the extracted features in a different context but still preserves the structure. Rotation has the major disadvantage that the reference depth is not defined for all pixels, which means that the uncertainty for these pixels cannot be evaluated. 
Random noise, on the other hand, is inconsistent across pixels, similar to diffusion-based augmentation. For random noise, the feature space augmentation results in improved outcomes compared to the image augmentation.
\begin{table}[!ht]
    \centering
   \caption{Uncertainty estimation result when applying different image augmentations or feature augmentations (marked by $^\star$) for the reference depth generation. The results are given for Monodepth2 (MD) and MonoViT (MV) trained on KITTI using monocular sequences, as well as for Monodepth2 (MD) trained on NYU.}
    \begin{tabular}{@{\hskip0pt}c@{\hskip1pt}@{\hskip2pt}l@{\hskip1pt}@{\hskip0pt}c@{\hskip1pt}@{\hskip0pt}c@{\hskip1pt}@{\hskip0pt}c@{\hskip1pt}@{\hskip0pt}c@{\hskip1pt}@{\hskip0pt}c@{\hskip1pt}@{\hskip0pt}c@{\hskip1pt}@{\hskip0pt}c@{\hskip0pt}}
        \toprule 
         & & \multicolumn{2}{c}{Abs Rel} & \multicolumn{2}{c}{RMSE} & \multicolumn{2}{c}{$\delta \geq 1.25$} \\
        \midrule 
        Setup & Loss & AUSE$^\downarrow$ & AURG$^\uparrow$ &  AUSE$^\downarrow$ & AURG$^\uparrow$ &  AUSE$^\downarrow$ & AURG$^\uparrow$ & nUCE$^\downarrow$ \\
        \midrule
       \multirow{10}{*}{\shortstack{KITTI\\MD\\Mono}} & \cellcolor{grays}Flip & \cellcolor{grays}0.029 & \cellcolor{grays}0.029 & \cellcolor{grays}0.533 & \cellcolor{grays}2.833 & \cellcolor{grays}\textbf{0.025} & \cellcolor{grays}\textbf{0.055} & \cellcolor{grays}0.005 \\
        & Gray & 0.029 & 0.029 & 0.531 & 2.836 & \textbf{0.025} & \textbf{0.055} & 0.005\\
        & Noise & 0.029 & 0.029 & 0.591 & 2.776 & 0.026 & 0.054 & 0.005 \\ 
        & Rot-5° & 0.031 & 0.027 & 0.679 & 2.687 & 0.029 & 0.051 & 0.008 \\
        & Rot-10° & 0.032 & 0.026 & 0.743 & 2.624 & 0.032 & 0.048 & 0.010 \\
        & Rot-20° & 0.033 & 0.025 & 0.721 & 2.646 & 0.033 & 0.047 & 0.013\\
        & Diff-R & 0.031 & 0.027 & 0.587 & 2.779 & 0.028 & 0.052 & 0.010 \\
        & Diff-S & 0.030 & 0.028 & 0.588 & 2.778 & 0.027 & 0.053 & 0.010 \\
        & \cellcolor{grays}Flip$^\star$ & \cellcolor{grays}\textbf{0.028} & \cellcolor{grays}\textbf{0.030} & \cellcolor{grays}0.542 & \cellcolor{grays}2.824 & \cellcolor{grays}0.026 & \cellcolor{grays}0.054 & \cellcolor{grays}\textbf{0.004} \\ 
        & Noise$^\star$ & 0.029 & 0.029 & \textbf{0.496} & \textbf{2.870} & \textbf{0.025} & \textbf{0.055} & \textbf{0.004} \\
        \midrule
        \multirow{10}{*}{\shortstack{KITTI\\MV\\Mono}}& \cellcolor{grays}Flip & \cellcolor{grays}0.028 & \cellcolor{grays}0.022 & \cellcolor{grays}0.657 & \cellcolor{grays}2.305 & \cellcolor{grays}0.022 & \cellcolor{grays}0.039 & \cellcolor{grays}\textbf{0.004} \\
        & Gray & 0.027 & 0.023 & 0.614 & 2.348 & \textbf{0.019} & 0.042 & 0.005 \\
        & Noise & 0.029 & 0.021 & 0.701 & 2.261 & 0.024 & 0.038 & 0.005 \\ 
        & Rot-5° & 0.027 & 0.023 & 0.547 & 2.415 & \textbf{0.019} & \textbf{0.043} & 0.007 \\
        & Rot-10° & 0.027 & 0.023 & 0.556 & 2.406 & \textbf{0.019} & \textbf{0.043} & 0.009 \\
        & Rot-20° & 0.027 & 0.023 & 0.555 & 2.407 & \textbf{0.019} & \textbf{0.043} & 0.011 \\
        & Diff-R & 0.034 & 0.016 & 0.831 & 2.132 & 0.030 & 0.032 & 0.006 \\
        & Diff-S & 0.033 & 0.017 & 0.795 & 2.167 & 0.028 & 0.034 & 0.005 \\
        & \cellcolor{grays}Flip$^\star$ & \cellcolor{grays}\textbf{0.026} & \cellcolor{grays}\textbf{0.024} & \cellcolor{grays}\textbf{0.512} & \cellcolor{grays}\textbf{2.450} & \cellcolor{grays}\textbf{0.019} & \cellcolor{grays}\textbf{0.043} & \cellcolor{grays}\textbf{0.004} \\ 
        & Noise$^\star$ & 0.028 & 0.022 & 0.635 & 2.327 & 0.021 & 0.040 & \textbf{0.004} \\
        \midrule 
        \multirow{9}{*}{\shortstack{NYU\\MD}}
        & GT & \textit{0.020} & \textit{0.066} & \textit{0.106} & \textit{0.265} & \textit{0.032} & \textit{0.123} & 0.120 \\
        & \cellcolor{grays}Flip & \cellcolor{grays}\textbf{0.062} & \cellcolor{grays}\textbf{0.025} & \cellcolor{grays}\textbf{0.252} & \cellcolor{grays}\textbf{0.119} & \cellcolor{grays}0.107 & \cellcolor{grays}0.048 & \cellcolor{grays}0.092 \\ 
        & Gray & 0.066 & 0.020 & 0.254 & 0.117 & 0.111 & 0.043 & 0.106 \\
        & Noise & 0.065 & 0.021 & 0.255 & 0.116 & 0.108 & 0.047 & 0.088 \\
        & Rot-5° & 0.067 & 0.019 & 0.268 & 0.103 & 0.115 & 0.040  & \textbf{0.051} \\
        & Rot-10° & 0.070 & 0.016 & 0.272 & 0.100 & 0.126 & 0.029 & 0.084 \\
        & Rot-20° & 0.077 & 0.010 & 0.271 & 0.100 & 0.145 & 0.010 & 0.162 \\
        & \cellcolor{grays}Flip$^\star$ & \cellcolor{grays}0.064 & \cellcolor{grays}0.022 & \cellcolor{grays}0.274 & \cellcolor{grays}0.098 & \cellcolor{grays}\textbf{0.106} & \cellcolor{grays}\textbf{0.049} & \cellcolor{grays}0.080 \\
        & Noise$^\star$ & 0.067 & 0.019 & 0.297 & 0.075 & 0.113 & 0.042 & 0.068 \\ 
        \bottomrule
    \end{tabular}
    \label{tab:ref_depth}
\end{table}
\subsubsection{Auxiliary Loss for Predictive Depth Estimation Models}
Table~\ref{tab:loss_importance} examines different choices as auxiliary loss functions for gradient extraction from predictive depth estimation models. Note that we conducted the ablation study using the \textit{Log-model}. We evaluate the options to use $\mathcal{L}_{aux,\sigma}$ (Eq.~\ref{eq:laux}), $\mathcal{L}_{aux}$ (Eq.~\ref{eq:lauxsigma}) without considering the predicted variance, and only the predicted variance as $\sigma^2$. In most of the metrics, the selected loss function  $\mathcal{L}_{aux,\sigma}$ achieves the best uncertainty estimation results. Observing the nUCE, the AUSE, and AURG in terms of RMSE, it is visible that only using the predicted variance $\sigma^2$ results in poor uncertainty estimates for the models trained on KITTI. In the case of NYU, taking the variance into account improves the uncertainty estimates. Therefore, combining the information from the consistency between the two depth predictions from the loss $\mathcal{L}_{aux}$ and the predicted variance $\mathbf{\sigma}$ to the auxiliary loss $\mathcal{L}_{aux,\sigma}$ for predictive depth estimation models results in the most stable uncertainty estimates. 
\begin{table}[!ht]
    \centering
   \caption{Uncertainty estimation results for different loss functions for gradient extraction on predictive depth estimation models. Results are given for Monodepth2 (MD) and MonoViT (MV) trained on KITTI using monocular sequences, and for Monodepth2 (MD) trained on NYU.}
    \begin{tabular}{@{\hskip0pt}c@{\hskip1pt}@{\hskip2pt}c@{\hskip1pt}@{\hskip0pt}c@{\hskip1pt}@{\hskip0pt}c@{\hskip1pt}@{\hskip0pt}c@{\hskip1pt}@{\hskip0pt}c@{\hskip1pt}@{\hskip0pt}c@{\hskip1pt}@{\hskip0pt}c@{\hskip1pt}@{\hskip0pt}c@{\hskip0pt}}
        \toprule 
         \multicolumn{2}{c}{} & \multicolumn{2}{c}{Abs Rel} & \multicolumn{2}{c}{RMSE} & \multicolumn{2}{c}{$\delta \geq 1.25$} & \\
        \midrule 
        Setup & Loss & AUSE$^\downarrow$ & AURG$^\uparrow$ &  AUSE$^\downarrow$ & AURG$^\uparrow$ & AUSE$^\downarrow$ & AURG$^\uparrow$ & nUCE$^\downarrow$ \\
        \midrule 
        \multirow{3}{*}{\makecell{KITTI\\MD\\Mono}} & \cellcolor{grays}$\mathcal{L}_{aux,\sigma}$ & \cellcolor{grays}\textbf{0.026} & \cellcolor{grays}\textbf{0.033} & \cellcolor{grays}0.824 & \cellcolor{grays}2.655 & \cellcolor{grays}\textbf{0.024} & \cellcolor{grays}\textbf{0.059} & \cellcolor{grays}0.006 \\
         & $\sigma^{2}$ & 0.036 & 0.023 & 2.399 & 1.080 & 0.042 & 0.041 & 0.119 \\
         & $\mathcal{L}_{aux}$ & 0.028 & 0.031 & \textbf{0.572} & \textbf{2.907} & 0.027 & 0.056 & \textbf{0.005} \\
        \midrule 
        \multirow{3}{*}{\makecell{KITTI\\MV\\Mono}} & \cellcolor{grays}$\mathcal{L}_{aux,\sigma}$ & \cellcolor{grays}\textbf{0.028} & \cellcolor{grays}\textbf{0.024} & \cellcolor{grays}0.870 & \cellcolor{grays}2.149 & \cellcolor{grays}\textbf{0.020} & \cellcolor{grays}\textbf{0.043} & \cellcolor{grays}0.004 \\
        & $\sigma^{2}$ & 0.038 & 0.014 & 2.391 & 0.628 & 0.038 & 0.025 & 0.229 \\ 
        & $\mathcal{L}_{aux}$ & 0.029 & 0.022 & \textbf{0.666} & \textbf{2.353} & 0.022 & 0.041 & \textbf{0.003} \\
        \midrule
         \multirow{3}{*}{\makecell{NYU\\MD}} & \cellcolor{grays}$\mathcal{L}_{aux,\sigma}$ & \cellcolor{grays}\textbf{0.054} & \cellcolor{grays}\textbf{0.031} & \cellcolor{grays}0.177 & \cellcolor{grays}0.190 & \cellcolor{grays}0.088 & \cellcolor{grays}0.067 & \cellcolor{grays}0.091 \\
         & $\sigma^{2}$ & \textbf{0.054} & \textbf{0.031} & \textbf{0.164} & \textbf{0.203} & \textbf{0.086} & \textbf{0.068} & \textbf{0.078} \\
         & $\mathcal{L}_{aux}$ & 0.059 & 0.026 & 0.236 & 0.131 & 0.100 & 0.054 & 0.115 \\
        \bottomrule 
    \end{tabular}
    \label{tab:loss_importance}
\end{table}
\subsubsection{Layer Selection}
The uncertainty estimation results when calculating the uncertainty score from a single decoder layer are reported for different decoder layers in Table~\ref{tab:layer_selection}. In general, we consider 9 layers for Monodepth2 and 8 layers for MonoViT, where we only consider convolutional layers that are not followed by an attention layer. Table~\ref{tab:layer_selection} lists the results of the last 5 layers. 
Overall, there is only a small difference between the uncertainty estimation results when using different layers. 
While the results in the case of Monodpeth2 are slightly better for earlier layers, the results in the case of MonoViT are slightly better for later layers for most metrics.
Therefore, it can be concluded that layer selection is architecture dependent, which can be avoided by using gradients from multiple layers as proposed in Section~\ref{sec:gradex}. Using the gradients from multiple layers removes the need to select a specific layer. 
\begin{table}[!ht]
    \centering
   \caption{Uncertainty estimation result for gradients extracted from different layers. The results are given for Monodepth2 (MD) and MonoViT (MV) trained on KITTI using monocular sequences, as well as for Monodepth2 (MD) trained on NYU.}
    \begin{tabular}{@{\hskip0pt}c@{\hskip1pt}@{\hskip2pt}c@{\hskip1pt}@{\hskip0pt}c@{\hskip1pt}@{\hskip0pt}c@{\hskip1pt}@{\hskip0pt}c@{\hskip1pt}@{\hskip0pt}c@{\hskip1pt}@{\hskip0pt}c@{\hskip1pt}@{\hskip0pt}c@{\hskip1pt}@{\hskip0pt}c@{\hskip0pt}}
        \toprule 
         & & \multicolumn{2}{c}{Abs Rel} & \multicolumn{2}{c}{RMSE} & \multicolumn{2}{c}{$\delta \geq 1.25$} \\
        \midrule 
         Setup & Layer & AUSE$^\downarrow$ & AURG$^\uparrow$ &  AUSE$^\downarrow$ & AURG$^\uparrow$ &  AUSE$^\downarrow$ & AURG$^\uparrow$ & nUCE$^\downarrow$\\
         \midrule 
         \multirow{5}{*}{\makecell{KITTI\\MD\\Mono}} & 5 & \textbf{0.028} & \textbf{0.030} & \textbf{0.514} & \textbf{2.852} & \textbf{0.025} & \textbf{0.055} & 0.005 \\
         & \cellcolor{grays}6 & \cellcolor{grays}0.029 & \cellcolor{grays}0.029 & \cellcolor{grays}0.533 & \cellcolor{grays}2.833 & \cellcolor{grays}\textbf{0.025} & \cellcolor{grays}\textbf{0.055} & \cellcolor{grays}0.005 \\
         & 7 & 0.029 & 0.029 & 0.573 & 2.793 & 0.026 & 0.054 & 0.004 \\
         & 8 & 0.029 & 0.029 & 0.618 & 2.749 & 0.027 & 0.053 & 0.004 \\
         & 9 & 0.030 & 0.028 & 0.637 & 2.729 & 0.027 & 0.053 & \textbf{0.003} \\
         \midrule
         \multirow{5}{*}{\makecell{KITTI\\MV\\Mono}} & 4 & 0.031 & 0.019 & 0.853 & 2.109 & 0.027 & 0.035 & \textbf{0.004} \\
         & \cellcolor{grays}5 & \cellcolor{grays}\textbf{0.028} & \cellcolor{grays}\textbf{0.022} & \cellcolor{grays}0.657 & \cellcolor{grays}2.305 & \cellcolor{grays}0.022 & \cellcolor{grays}0.039 & \cellcolor{grays}\textbf{0.004} \\
         & 6 & 0.032 & 0.018 & 0.851 & 2.111 & 0.028 & 0.034 & \textbf{0.004} \\
         & 7 & 0.028 & 0.022 & \textbf{0.483} & \textbf{2.479} & \textbf{0.019} & \textbf{0.043} & \textbf{0.004} \\
         & 8 & 0.028 & 0.022 & 0.510 & 2.452 & \textbf{0.019} & \textbf{0.043} & \textbf{0.004} \\
        \midrule
         \multirow{5}{*}{\makecell{NYU\\MD}} & 5 & 0.063 & 0.023 & \textbf{0.237} & \textbf{0.134} & 0.112 & 0.043 & 0.094 \\
         & \cellcolor{grays}6 & \cellcolor{grays}\textbf{0.062} & \cellcolor{grays}\textbf{0.025} & \cellcolor{grays}0.252 & \cellcolor{grays}0.119 & \cellcolor{grays}\textbf{0.107} & \cellcolor{grays}\textbf{0.048} & \cellcolor{grays}0.092 \\
         & 7 & 0.064 & 0.022 & 0.259 & 0.112 & 0.111 & 0.043 & 0.092 \\
         & 8 & 0.065 & 0.021 & 0.284 & 0.087 & 0.115 & 0.040 & 0.087 \\
         & 9 & 0.066 & 0.020 & 0.298 & 0.073 & 0.118 & 0.037 & \textbf{0.075} \\
        \bottomrule 
    \end{tabular}
    \label{tab:layer_selection}
\end{table}
\subsubsection{Combining Gradients from Multiple Layers}
Instead of calculating the uncertainty score from a single layer, we introduce the option to combine the uncertainty maps from multiple layers. In our method, we use the maximum (MAX) over the layers as a combination function. In Table~\ref{tab:score_selection}, we also assess the mean (MEAN) and the variance (VAR) over the uncertainty maps as alternatives. In most cases, the best results are obtained with MAX, while MEAN also gives good results in uncertainty estimation. 
\begin{table}[!ht]
    \centering
   \caption{Uncertainty estimation result when using gradients from multiple layers, combining the uncertainty maps from each layer with the mean (MEAN), variance (VAR), or, the maximum (MAX). The results are given for Monodepth2 (MD) and MonoViT (MV) trained on KITTI using monocular sequences, as well as for Monodepth2 (MD) trained on NYU.}
    \begin{tabular}{@{\hskip0pt}c@{\hskip1pt}@{\hskip2pt}c@{\hskip1pt}@{\hskip0pt}c@{\hskip1pt}@{\hskip0pt}c@{\hskip1pt}@{\hskip0pt}c@{\hskip1pt}@{\hskip0pt}c@{\hskip1pt}@{\hskip0pt}c@{\hskip1pt}@{\hskip0pt}c@{\hskip1pt}@{\hskip0pt}c@{\hskip0pt}}
        \toprule 
         & & \multicolumn{2}{c}{Abs Rel} & \multicolumn{2}{c}{RMSE} & \multicolumn{2}{c}{$\delta \geq 1.25$} \\
        \midrule 
         Setup & Score & AUSE$^\downarrow$ & AURG$^\uparrow$ &  AUSE$^\downarrow$ & AURG$^\uparrow$ &  AUSE$^\downarrow$ & AURG$^\uparrow$ & nUCE$^\downarrow$ \\
         \midrule 
         \multirow{3}{*}{\makecell{KITTI\\MD\\Mono}} & MEAN & \textbf{0.029} & \textbf{0.029} & 0.562 & 2.805 & 0.026 & 0.054 & \textbf{0.004} \\
         & VAR & \textbf{0.029} & \textbf{0.029} & 0.553 & 2.814 & 0.026 & 0.054 & \textbf{0.004} \\
         & \cellcolor{grays}MAX & \cellcolor{grays}\textbf{0.029} & \cellcolor{grays}\textbf{0.029} & \cellcolor{grays}\textbf{0.548} & \cellcolor{grays}\textbf{2.819} & \cellcolor{grays}0.026 & \cellcolor{grays}0.054 & \cellcolor{grays}0.005 \\
         \midrule
         \multirow{3}{*}{\makecell{KITTI\\MV\\Mono}} & MEAN  & \textbf{0.026} & \textbf{0.024} & \textbf{0.484} & \textbf{2.478} & \textbf{0.018} & \textbf{0.044} & \textbf{0.004} \\
         & VAR & 0.030 & 0.020 & 0.709 & 2.253 & 0.024 & 0.037 & 0.005 \\
         & \cellcolor{grays}MAX & \cellcolor{grays}0.027 & \cellcolor{grays}0.023 & \cellcolor{grays}0.549 & \cellcolor{grays}2.413 & \cellcolor{grays}0.019 & \cellcolor{grays}0.042 & \cellcolor{grays}0.005 \\
        \midrule
         \multirow{3}{*}{\makecell{NYU\\MD}} & MEAN & 0.063 & \textbf{0.024} & 0.262 & 0.109 & 0.109 & 0.046 & 0.087 \\
         & VAR & \textbf{0.062} & \textbf{0.024} & 0.257 & 0.114 & \textbf{0.105} & \textbf{0.050} & \textbf{0.063} \\
         & \cellcolor{grays}MAX & \cellcolor{grays}\textbf{0.062} & \cellcolor{grays}\textbf{0.024} & \cellcolor{grays}\textbf{0.256} & \cellcolor{grays}\textbf{0.115} & \cellcolor{grays}0.107 & \cellcolor{grays}0.048 & \cellcolor{grays}0.122 \\
        \bottomrule 
    \end{tabular}
    \label{tab:score_selection}
\end{table}

\section{Conclusion}
We introduce a novel \textit{post hoc} uncertainty estimation approach based on gradients extracted from an already trained depth estimation model. To obtain meaningful gradients without relying on ground truth depth, we define an auxiliary loss function as the difference between the predicted depth and a reference depth. The reference depth, which acts as pseudo ground truth, is generated using a simple image or feature augmentation, making our method simple and effective. The derivatives \textit{w.r.t.} the feature maps are computed via back-propagation to determine the uncertainty score from gradients extracted from a single layer or multiple layers. We show detailed insights into the use of gradients for uncertainty estimation by evaluating the key characteristics of our approach. Our experiments on KITTI and NYU show the effectiveness of our gradient-based approach to determine the uncertainty of depth estimation models without the need for re-training. In particular, for models trained with monocular sequences, which are most prone to uncertainty, our method dominates in all metrics, including the proposed nUCE.

\ifCLASSOPTIONcompsoc
  \section*{Acknowledgments}
  Part of the research leading to these results is funded by the German Federal Ministry for Economic Affairs and Climate Action" within the project “KI Delta Learning“ (Förderkennzeichen 19A19013A). The authors would like to thank the consortium for the successful cooperation. Also, this work was partially funded by Deutsche Forschungsgemeinschaft (DFG) through the project "Transferring Deep Neural Networks from Simulation to Real-World" (project number 458972748).
  
\else
  \section*{Acknowledgment}
\fi
\ifCLASSOPTIONcaptionsoff
  \newpage
\fi



%
\bibliography{tpami_2023}
\bibliographystyle{IEEEtran}

%

\begin{IEEEbiography}[{\includegraphics[width=1in,height=1.25in,clip,keepaspectratio]{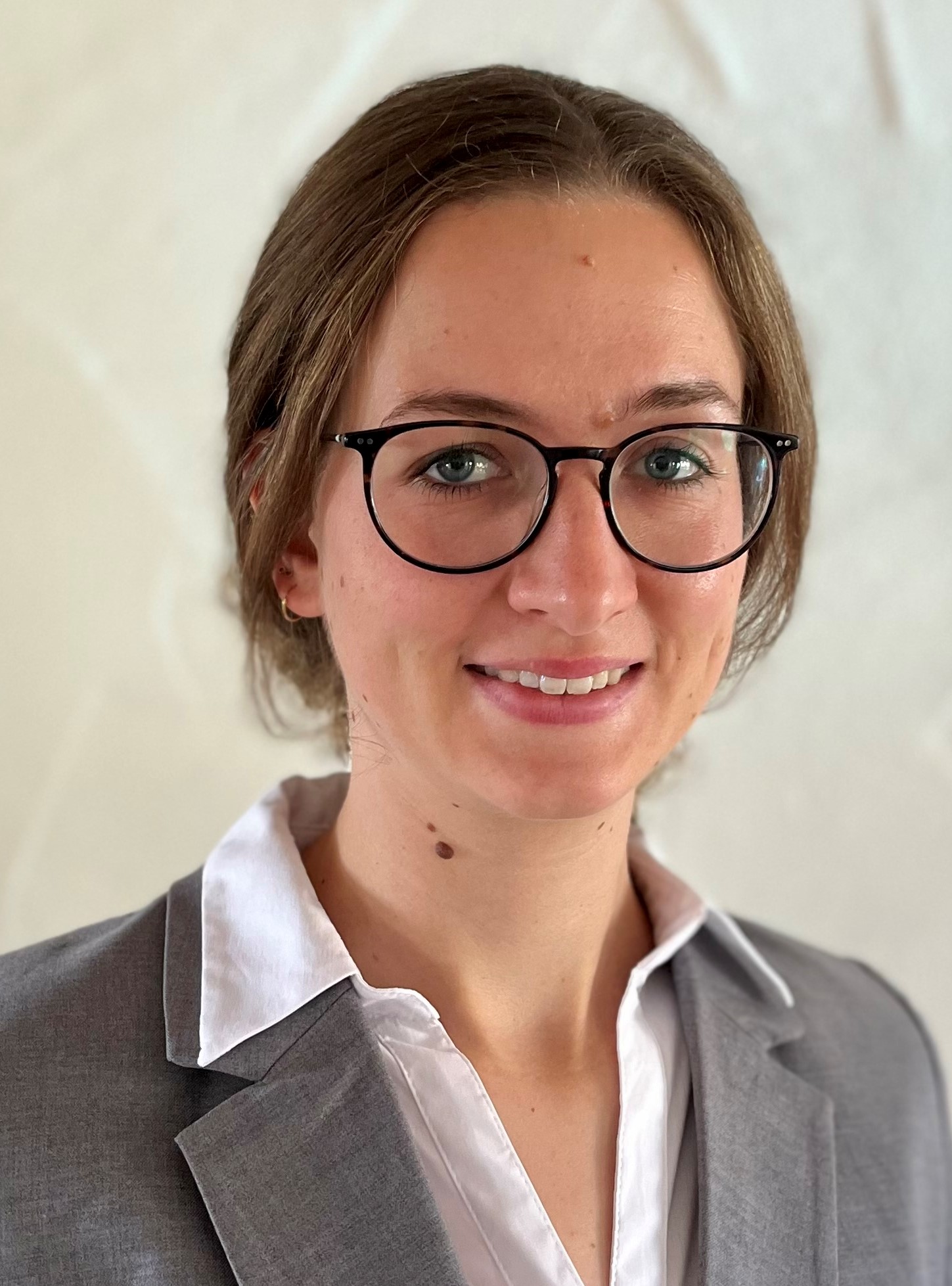}}]{Julia Hornauer} received her doctorate in 2024 from Ulm University, Germany, at the institute for Measurement, Control and Microtechnology. Her main research interests are computer vision for autonomous driving. The focus of her work is on uncertainty estimation and out-of-distribution detection for monocular depth estimation. She holds a Master's degree in Communications and Computer Engineering from Ulm University, Germany.

\end{IEEEbiography}
\begin{IEEEbiography}[{\includegraphics[width=1in,height=1.25in,clip,keepaspectratio]{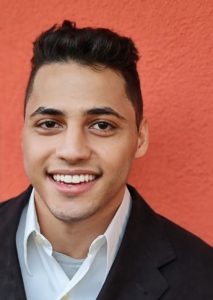}}]{Amir El-Ghoussani}
holds a Master's degree in Medical Engineering from the Faculty of Engineering at the Friedrich-Alexander-Universität Erlangen-Nürnberg. Since 2021 he has started pursuing a PhD degree at Friedrich-Alexander-Universität Erlangen-Nürnberg at the Chair of Multimedia Communications and Signal Processing. His research primarily focuses on computer vision and deep learning, with a particular interest in exploring their applications in sim2real approaches.
\end{IEEEbiography}
\begin{IEEEbiography}[{\includegraphics[width=1in,height=1.25in,clip,keepaspectratio]{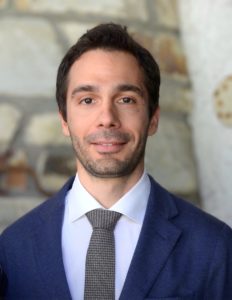}}]{Vasileios Belagiannis} is a professor at the Faculty of Engineering at the Friedrich-Alexander-Universität Erlangen-Nürnberg. He holds a degree in Engineering from the Democritus University of Thrace, Engineering School of Xanthi and M.Sc. in Computational Science and Engineering from TU München. He completed his doctoral studies at TU München and then continued as post-doctoral researcher at the University of Oxford (Visual Geometry Group). His research interests include representation learning, uncertainty estimation, anomaly detection, multi-modal learning, learning algorithm for noisy labels, few-shot learning and hardware-aware learning.
\end{IEEEbiography}
\vfill
\end{document}